\documentclass[sigconf]{acmart}
\renewcommand\footnotetextcopyrightpermission[1]{} 

\usepackage{booktabs}

\usepackage{url,bm}
\usepackage{latexsym}

\usepackage{etoolbox}
\newtoggle{conf}
\toggletrue{conf}

\newcommand{\cut}[1]{}

\newenvironment{compact_enum}{
\begin{itemize}
  \setlength{\itemsep}{0pt}
  \setlength{\parskip}{1pt}
  \setlength{\parsep}{0pt}
  \setlength{\itemindent}{-5pt}
}{\end{itemize}}

\newenvironment{compact_numbered_enum}{
\begin{enumerate}
  \setlength{\itemsep}{0pt}
  \setlength{\parskip}{0pt}
  \setlength{\parsep}{0pt}
  \setlength{\itemindent}{-5pt}
}{\end{enumerate}}

\usepackage{subfigure}
\usepackage{tikz}
\usetikzlibrary{arrows,shapes,snakes,automata,backgrounds,petri}
\usepackage[latin1]{inputenc}
\usepackage{verbatim}
\usepackage[ruled,vlined,linesnumbered,commentsnumbered]{algorithm2e}

\usepackage{url}
\usepackage{verbatim}
\usepackage{multirow}
\usepackage{amsmath}

\DeclareMathOperator*{\argmax}{arg\,max}

\begin{document}
\title{In-Session Personalization for Talent Search}

\author{Sahin~Cem~Geyik, Vijay~Dialani, Meng~Meng, Ryan~Smith}
\affiliation{%
  \institution{LinkedIn Corporation}
}

\renewcommand{\shortauthors}{S. C. Geyik et al.}

\begin{abstract}

Previous efforts in recommendation of candidates for talent search followed the general pattern of receiving an initial search criteria and generating a set of candidates utilizing a pre-trained model. Traditionally, the generated recommendations are final, that is, the list of potential candidates is not modified unless the user explicitly changes his/her search criteria. In this paper, we are proposing a candidate recommendation model which takes into account the immediate feedback of the user, and updates the candidate recommendations at each step. This setting also allows for very uninformative initial search queries, since we pinpoint the user's intent due to the feedback during the search session. To achieve our goal, we employ an intent clustering method based on topic modeling which separates the candidate space into meaningful, possibly overlapping, subsets (which we call \emph{intent clusters}) for each position. On top of the candidate segments, we apply a multi-armed bandit approach to choose which intent cluster is more appropriate for the current session. We also present an online learning scheme which updates the intent clusters within the session, due to user feedback, to achieve further personalization. Our offline experiments as well as the results from the online deployment of our solution demonstrate the benefits of our proposed methodology.
\end{abstract}


\maketitle

\section{Introduction}
\label{sec:introduction}
LinkedIn is the largest professional network on the web which connects more than 500 million professionals world-wide. It also has become the primary source for corporations around the world to find new talent for their specific needs. Indeed, around $65\%$ of the company's revenue is due to the Talent Solutions\footnote{~ https://press.linkedin.com/about-linkedin \\ ~ \\ ~  \large \textbf{This paper has been accepted for publication at ACM CIKM 2018.}} products, which are tailored towards presenting our customers with the most beneficial future employees.

Efforts for relevance in talent search applications have so far employed offline generated models utilized to rank potential candidates, and present the user (a recruiter or a hiring manager) a static list of recommendations. This ranking depends on the initial input of the user (that is utilized by the model), which is often a carefully crafted set of search terms \cite{hathuc2015expertisesearch}. The biggest problem with such an approach is the fact that it requires deep domain knowledge from the user, as well as significant time and manual effort to come up with the best search criteria (e.g. which skills are relevant for a specific role that the recruiter is looking to fill). \cite{hathuc2016talentsearch} addresses this problem by allowing a user to list a multitude of \emph{ideal candidates}. These candidates can then be translated into an implicit and focused search query.

A shortcoming of the previous approaches is the fact that as the user examines the recommended candidates and gives feedback\footnote{~Feedback in general is an action performed by the user, such as clicking on, connecting with, messaging, or rating a recommended candidate. In the context of the current work, we utilize the ratings of a user (two types, \emph{positive} (the candidate fits the requirements) and \emph{negative} (candidate does not fit the requirements)) to each recommended candidate, which are presented to the user one-by-one, in an online manner, i.e. each new recommended candidate depends on the feedback on the previous candidate(s).}, these are not taken into account during the current search session\footnote{~A search session encapsulates the recommended candidates to and feedback actions performed by a recruiter within the context of a single search query, i.e. without any modification to the initial search query.}. While each feedback is eventually included in the offline generated models, there is a significant opportunity loss by not immediately incorporating them for the current search results.

One significant advantage of improving the candidate quality due to immediate user feedback would be to further simplify the initial user input. In certain cases, even providing an ideal candidate may not be easy for a user, and instead the user may just want to give an initial uninformative query (e.g. the user could just provide a position to be filled, such as \emph{software engineer}). In such a case, a smart system should adapt to the user's feedback and after some steps, i.e. immediate feedback given to candidates that are presented one at a time, recommend the best candidates for the job. 

In this paper we focus on the above premise: Given an uninformative initial query, which we assume to be the \emph{position to be filled} in the context of this paper, we aim to improve the candidate quality step-by-step according to the immediate user feedback to the recommended candidates (i.e. without modification to the initial query, hence each step constitutes a single candidate recommended for the current query, and the model is updated by the rating feedback, given to each candidate, presented one-by-one). We first construct a meaningful segmentation of the candidate space (for each position, i.e. title) in an offline manner. Then, the per-title constructed segments are utilized by a multi-armed bandit (MAB) approach for selecting the most appropriate segment for the current search session in real-time, due to immediate feedback. Furthermore, we also propose to improve each offline generated segment's inner quality via an online learning approach applied per session.

The contributions of our work can be listed as:
\vspace{-0.05in}
\begin{compact_enum}
\item Utilization of topic models for offline segmentation of the candidate space into meaningful clusters for each position/title,
\item A model selection approach utilizing multi-armed bandits to choose the most appropriate candidate segment for the current session in real-time, and,
\item An online learning approach for improving candidate segments during the session due to user feedback to each candidate, presented one at a time.
\end{compact_enum}
The rest of the paper is organized as follows. In the next section, we give a formal definition of our problem, and the challenges in solving it. Later, we will present our proposed methodology, as well as the implementation details in \S~\ref{sec:methodology} and \S~\ref{sec:implementation}, respectively. Our empirical results follow in \S~\ref{sec:results}, along with the relevant work from the recommendation systems literature in \S~\ref{sec:related_work}. Finally, \S~\ref{sec:conclusions} concludes the paper.

\section{Problem Statement}
\label{sec:problem_def}

As mentioned before, our focus in this work is to take into account the user feedback to the recommended candidates for talent search, so that the candidate quality can be improved dynamically within a session. In recommender systems domain, the total utility of the items suggested by a recommendation algorithm is often evaluated by \emph{Discounted Cumulative Gain} metric \cite{jarvelin_2002}, which can be formulated as:
\begin{equation}
\label{eq:dcg}
\textrm{DCG}(S) = \sum_{t = 1}^{|S|} g(\textrm{q}(item_t), t) \textrm{ , where,}
\end{equation}
\[
g(q, t) = \frac {2^q - 1}{\log_2(t +1)} ~.
\]
In the above formulation $S$ is an ordered set of items, \emph{q} is a quality score given to an $item_t$ by the user of the recommendation system (in our case, items are candidates, and quality is a binary variable where 1/0 indicates whether the user liked the candidate or not), and $g$ is the \emph{gain} function which estimates the utility of this candidate (which is calculated from the quality of the candidate, and the order with which this candidate was shown)
\footnote{~ When comparing the performance of the same algorithm over different item sets, we need a normalized metric (since DCG is highly affected by the size of the set and total number of possible good items in it). For this purpose, mostly the \emph{Normalized} DCG metric is used, which can be formulated as:
\begin{equation}
\label{eq:ndcg}
\textrm{\emph{NDCG}} = \frac{\textrm{\emph{DCG}}}{\textrm{\emph{IDCG}}} ~ ,
\end{equation}
where \emph{IDCG} stands for \emph{Ideal} DCG, i.e. the best possible ranking of the set of items. In our case, IDCG is the DCG of an artificially re-ranked version of the item set where the highest quality items are at the earliest indices.}.
It follows from Eq.~\ref{eq:dcg} that a recommendation algorithm's aim should be to get the relevant (high quality) items as highly ranked as possible (i.e. higher quality score items with smaller denominators). Please note that any formulation that motivates the placement of the relevant items higher in the recommendation order can also be utilized for evaluation and improvement of recommender systems. In our case, a metric such as \emph{precision@k} (i.e. number of relevant items in the first k recommendations) is a more stable business metric, hence is what we utilize for our evaluations in \S~\ref{sec:results}. It is trivial to show that an algorithm that optimizes DCG also optimizes precision@k, though not necessarily the other way round.

Following the above discussions, next we formally describe our problem, which we term as \emph{In-Session Optimization} of Candidate Quality for Talent Search.
\begin{definition}
\label{def:prob_def}
Given the set of all possible candidates $S$ (a list indexed as $S[i]$, e.g. for $i^{th} element$, and $S[1:|S|]$ represents the whole set) that may be a potential fit to a talent search effort), and where,
\begin{compact_enum}
\item $S'_t$ represents an ordering of S at time-step t,
\item $P$ is a set of available policies $P_i$, where each $P_i$ dictates how $S$ is re-ranked at time-step t from $S'_{t-1}$ into $S'_t$ due to feedback to the $(t-1)^{th}$ candidate ($q(S'_{t-1}[t-1])$, which is binary as given in Eq.~\ref{eq:dcg}),
\end{compact_enum}
\emph{In-Session Optimization} aims to find an optimal policy $P_{opt}$ (selected amongst $P_i \in P$) such that:
\begin{equation}
\label{eq:p_opt}
P_{opt} = \argmax_{P_i \in P} \sum_{t = 1}^{|S|} g(\textrm{q}(S'_t[t]), t) \textrm{ , s.t.,} ~~~~~~~~~~~~~~~~~~~~~~~~~~~~~~~~~~~~~~~~
\end{equation}
\[
S'_t = P_i(S'_{t-1},~q(S'_{t-1}[t-1]))\textrm{~, and,} ~~~~~~~~~~~
\]
\[
S'_t[1:t-1] = S'_{t-1}[1:t-1] ~~~ \textrm{for any } t. ~~~~
\]
\end{definition}
The above definition describes a dynamic re-ranking of the potential candidate set at each time-step t (t goes from 1 to |S| since we can at most recommend |S| candidates, due to S being the set of candidates that fit the search criteria). The formulation explicitly states that at time-step (t-1), we show the candidate in $(t-1)^{th}$ index of the ranked candidate list $S'_{t-1}$ (i.e. $S'_{t-1}[t-1]$), get the feedback for that candidate ($q(S'_{t-1}[t-1])$), and update the ranking of candidates (using policy $P_i$), i.e. transform $S'_{t-1}$ into $S'_{t}$, which is a new ordering on the potential candidates. Since the candidates at indices $1$ through $t-1$ have already been shown, the formulation also states that $S'_t[1:t-1] = S'_{t-1}[1:t-1]$, i.e. we can only re-rank the candidates that have not yet been recommended to the user. Although the function we are optimizing in the definition is the DCG of the candidates shown at each step (i.e. $g(q, s)$ from Eq.~\ref{eq:dcg}), again, this can be replaced by any other metric that values the relevant candidates more in the earlier ranks. The optimal $P_{opt}$ decides both the first ranking of the set $S$, i.e. $S'_{1}$, as well as how to update the candidate ranking at each time-step (i.e. via the constraint $S'_t = P_i(S'_{t-1},~q(S'_{t-1}[t-1]))$ in the definition), by taking into account the feedback for the latest recommended candidate.

The challenges in a good choice of $P_{opt}$ are as follows:
\begin{compact_enum}
\item Initial choice of the ranking function which will present us with $S'_1$. This can be either a model trained offline over many users, or one which is highly specialized for the current user and his/her aim in candidate search. In our context, we constrain ourselves to a case where the initial information provided to us by the user is fairly uninformative (as mentioned before, the motivation is to enable people with little recruiting experience to be able to search for candidates). Due to this fact, the initial ranking should be as general as possible to explore potential user interests, but with a capability to improve as we get feedback. Ideally, we should have a high variety of candidates as early as possible in the ranks, so that the user can evaluate these different types of candidates and our policy can better understand the user's preferences.
\item The method in improving the ranking function due to the user's feedback (to the recommended candidate) at each step, where a step is defined as the process of showing a single candidate and receiving feedback (in terms of good fit or bad fit, similar to a thumbs up or thumbs down rating) from the user. If we are using a parametric model as our initial ranking algorithm, $P_{opt}$ should be able to update the parameters in a meaningful way. The choice of this method is crucial to the overall recommendation quality, since the initial model does not have specific enough information for the current session, i.e. the precise type of candidate the user deems appropriate for the current position s/he would like to fill.
\item A good balance between \emph{exploration vs. exploitation} which is a common challenge in recommender systems. In our context, as we learn from the feedback of the user, and bring more and more relevant candidates higher in the ranking (exploitation), we should leave opportunities to learn from the candidates that are not immediately useful, but may help us in further improving our ranking (exploration).
\end{compact_enum}
Based on the problem definition and the challenges associated with it, we will present the details of our recommendation scheme in the next section.

\section{Methodology}
\label{sec:methodology}

As presented in Definition~\ref{def:prob_def}, we aim to solve the \emph{in-session optimization} problem for talent search. Our use case assumes minimal input from the user at the beginning, e.g. a position to be filled, hence the potential candidates that would fit into such an uninformative initial query are too many and highly variant in their properties (skills, seniority etc.). Furthermore, we do not have a holistic view of the feedback that the user would give to all the possible candidates (i.e. we need to first pinpoint the user intent, hence direct optimization of Eq.~\ref{eq:p_opt} is unfeasible), though the overall aim is to be able to bring as many relevant candidates to the higher ranks (from Eq.~\ref{eq:p_opt}) by \emph{exploiting} the preference information received from user feedback to the recommended candidates, while \emph{exploring} the different types of candidates that the user may be interested in. To achieve these goals, we are proposing a candidate recommendation strategy as follows:
\begin{enumerate}
\item Segment the candidate space into meaningful groups, where each group constitutes a smaller search space,
\item Pinpoint which segment is the most appropriate for the user's current aim, and,
\item Update the segment definitions further due to the user's feedback to recommended candidates.
\end{enumerate}
The rest of this section details our methodology which executes on the above three points.

\subsection{Topic Modeling for \\Candidate Space Segmentation}
\label{subsec:lda_clus}
Each candidate that could potentially be a fit to the position to be filled can be taken as a bag of skills, seniority, previous companies, previous positions etc. (i.e. \emph{candidate properties}).  We propose to utilize topic modeling to separate the potential candidates into candidate topics per position, where each topic is a distribution over the candidate properties, and provides a means to do soft clustering (i.e. with overlapping candidates) to segment the candidate space. We call these soft clusters of candidates for each position as \emph{intent clusters}, i.e. a cluster of candidates for a specific intent that the user may have for the current session. To generate the intent clusters via topic modeling, we are applying Latent Dirichlet Allocation \cite{blei_2003_lda} (LDA) in our current work.

\subsubsection{Latent Dirichlet Allocation for Intent Clusters}
\label{subsubsec:lda_detailed}

Originally applied on text modeling, LDA assumes the following generative process for each document \textbf{d} in a corpus \textbf{D}:
\begin{compact_numbered_enum}
  \item Choose the number of words $\textit{N} \sim Poisson(\xi)$.
  \item Choose the multinomial topic distribution for d, i.e. $\theta \sim Dir(\alpha)$.
  \item For each word $w_n$ within $d$:
  \begin{compact_numbered_enum}
     \item Choose a topic $z_n \sim Multinomial(\theta)$, and,
     \item Choose a word $w_n$ from $p(w_n | z_n,  \beta)$, a multinomial probability conditioned on the topic $z_n$.
   \end{compact_numbered_enum}
\end{compact_numbered_enum}
Therefore, each document $d$ is a sequence of N words denoted by $w_{1 \rightarrow N}$, and the corpus $D$ is defined as a collection of $\textit{M}$ documents $d_{1 \rightarrow M}$, with a probability of:
\begin{equation}
\textrm{\scriptsize{$p(D | \alpha, \beta) = \prod_{d=1}^M \int_{\theta_d} p(\theta_d | \alpha) \left( \prod_{n=1}^{N_d}\sum_{z_{dn}} p(z_{dn} | \theta_d)p(w_{dn} | z_{dn}, \beta)\right)\mathrm{d}\theta_d ~.$}}
\label{eq:lda_prob}
\end{equation}
LDA is a probabilistic graphical model with a three level representation as given in Figure~1. The outer plate represents documents, and the inner plate represents the repeated choice of topics and words within a document. As indicated in Eq.~\ref{eq:lda_prob}, $\alpha$ and $\beta$ are corpus level parameters sampled once in the process of corpus generation. The variables $\theta_{d}$ are document level variables, sampled once per document. Finally, the variables $z_{dn}$ and $w_{dn}$ are word-level variables and are sampled once for each word in each document. Exact inference of LDA parameters are intractable \cite{blei_2003_lda}. Instead, variational inference \cite{blei_2003_lda} (which we also utilize for this work) and Markov-chain Monte Carlo \cite{Jordan1999} methods are often used for approximate inference of LDA parameters.

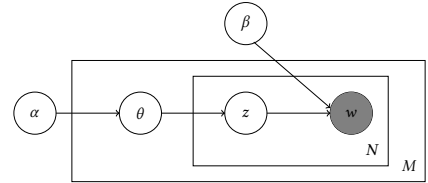
\begin{figure}[htb]
\label{fig:lda_plate}
\centering
\resizebox{2.2in}{!}{
\begin{tikzpicture}
\draw(0.8, 2.2) circle(0.4);
 \node at (0.8, 2.2) {$\beta$};
\draw (-2.5, 1.5) rectangle (4.2, -0.8);
 \node at (3.9, -0.5) {$M$};
\draw (-0.2, 1.2) rectangle (3.5, -0.5);
 \node at (3.2, -0.2) {$N$};
 \draw (-3.2, 0.5) circle (0.4);
  \draw (-1.2, 0.5) circle (0.4);
  \draw (0.8, 0.5) circle (0.4);
    \draw (2.8, 0.5)  circle (0.4);
    \fill[color=gray]  (2.8, 0.5)  circle (0.4);
 \draw[->](-2.8,0.5)--(-1.6,0.5);
  \draw[->](-0.8,0.5)--(0.4,0.5);
   \draw[->](1.2,0.5)--(2.4,0.5);
     \draw[->](0.95,1.85)--(2.4,0.6);
 \node at (-3.2, 0.5) {$\alpha$};
  \node at (-1.2, 0.5) {$\theta$};
   \node at (0.8, 0.5) {$z$};
    \node at (2.8, 0.5) {$w$};
\end{tikzpicture}
}
\vspace{-10pt}
\caption{Graphical model representation of LDA.}
\end{figure}

In our application, each potential candidate is a document, the words within the document are the skills and seniority tokens extracted from that candidate's profile, and the topics generated are the intent clusters which represent a distribution over skills, seniority etc., i.e. \emph{candidate properties}. Table~\ref{table:example_clusters} presents an example set of clusters (topics, where we removed the probabilities for presentational purposes) with the skills for the position \emph{Software Engineer}, generated  from the candidate corpus at LinkedIn (utilizing those members that are Software Engineers as the documents). The last column is our manual interpretation of the cluster.

\begin{table}
\centering
\smaller
\caption{Example Clusters for Software Engineering}
\vspace{-4pt}
\begin{tabular}{|c|c|c|} \hline
Id & Skills & Interpretation \\ \hline
1	& ajax, java, spring framework, 		& J2EE \\
	& android, javascript, xml 				& Developer \\ \hline
2	& php, javascript, html5,					& Front-end \\
	& ajax, css, html 							& Developer  \\ \hline
3	& java, matlab, git, 							& Unix/Linux \\
	& python, linux, unix 						& Developer \\ \hline
4	& android development, sql,			& Android \\
	& mysql, linux, php, html 				& Developer  \\ \hline
\end{tabular}
\label{table:example_clusters}
\end{table}

\begin{figure}[ht]
\centering
\includegraphics[width=2.1in]{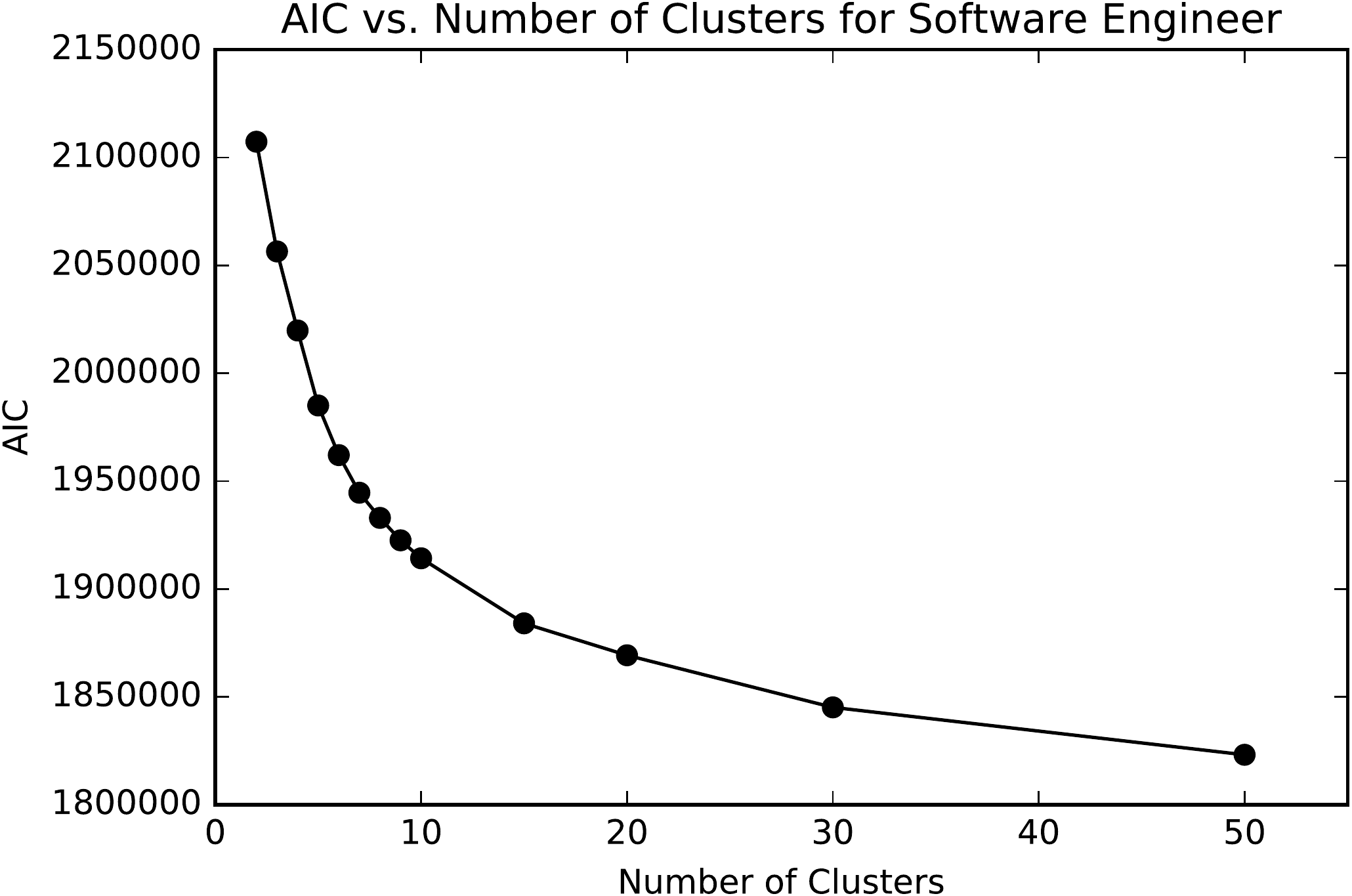}
\vspace{-10pt}
\caption{AIC vs. Number of Clusters for Software Engineer}
\label{fig:aic_swe}
\end{figure}

A special care often needs to be given to the choice of the number of topics, taking into account how the selected value effects the quality of separation. As an example, Figure~\ref{fig:aic_swe} shows the \emph{Akaike Information Criterion} (AIC)\footnote{~ AIC estimates the loss of information in representing the process that originally generated the documents when we set a model variable (in this case the number of topics), hence the lower this metric, the better.} \cite{akaike_1974} results of the effect of changing number of clusters for the position \emph{Software Engineer}. From the figure, it seems that the rate of drop for AIC seems to stall starting with around five clusters, which is a potentially good number of intent clusters to be utilized for segmenting software engineers (from our experiments, this value changes between 5-10 for other positions as well).

\subsubsection{Recommendation via Intent Clusters}
\label{subsubsec:utilizing_intent_clusters}
While the intent clusters for each position/title are generated and stored offline as a distribution over candidate properties, we need a way to ensure a ranked flow of candidates from each cluster at recommendation time. We propose to utilize each intent cluster\footnote{At the time of recommending a next candidate, based on the title that the user had entered as a search query for the current session, we pick up the stored set of intent clusters for that title and serve candidates from them. We present the details of how we choose the next intent cluster to serve a candidate from at each step in the next section.}, and therefore the user properties that they represent a distribution over, as a meta-query to hit the in-house search engine at LinkedIn, called \emph{Galene} \cite{galene_engine}. Galene allows for arbitrary ranking functions to be applied to the set of candidates that match a specific query, and we are utilizing a ranking function similar to the one given in \cite{hathuc2015expertisesearch}, which is trained offline.

Even after applying a ranking function which is trained on a general set of users utilizing offline data, it is not guaranteed that we will take into account the distribution over candidate properties for each intent cluster which is the output of our topic modeling. Hence, we further personalize the ranking of the candidates returned by our search engine with a linear combination of each candidate and the intent cluster's distribution. Therefore, each candidate $c_m$ can be scored by each intent cluster $t_n$ for further ranking as:
\begin{equation}
\label{eq:linear_score}
matchScore(c_m | t_n) = \vec{w}_{t_n} \bullet \vec{w}_{c_m} = \sum_i w_{t_n,i} \cdot w_{c_m,i}  ~,
\end{equation}
where $\vec{w}_{c_m}$ is a binary vector of whether the candidate $c_m$ has a specific property (skill, seniority, position etc.), and $\vec{w}_{t_n}$ is a vector representing the distribution of the intent cluster $t_n$ over possible candidate properties. This formulation is equivalent to taking a weighted sum of the intersection (of properties) between the candidate and the intent cluster, therefore it measures the similarity (hence the name \emph{matchScore}) between the candidate returned by an intent cluster, and the cluster itself (higher the similarity, higher the rank). After the offline ranking score  \cite{hathuc2015expertisesearch} (which one may call \emph{offlineScore}) and \emph{matchScore} are calculated per candidate ($c_m$), our final ranking score takes a convex combination of the two:
\begin{equation}
\label{eq:convex_score}
score(c_m | t_n) = \alpha ~ matchScore(c_m | t_n) ~ + ~ (1 - \alpha) ~ \textrm{\emph{offlineScore}}(c_m | t_n) ~ ,
\end{equation}
and return the candidates in the descending order of their scores. We evaluate the choice of $\alpha$ and how it affects the recommendation quality in \S~\ref{sec:results}.

Since any candidate may have properties that span over (intersect with) different intent clusters, it is possible that the same candidate can appear in the ranking of multiple intent clusters. However, the similarity score helps with getting the candidates with the highest match to the top, hence improving the distinctness of the intent clusters especially in the earlier ranks.

\subsection{Multi-Armed Bandits for \\Intent Cluster Selection}
\label{subsec:mab_clus}
The next step in our recommendation scheme is understanding the user's intent in the current session, i.e. which intent cluster/s the user is most inclined with. The intent clusters do help us in reducing the space of candidates to recommend; on the other hand, choosing the best intent cluster is an \emph{algorithm selection} problem \cite{kotthoff_2014}, and is also closely tied with \emph{meta-learning} concept in Machine Learning \cite{aha_1992, miles_2008}.

In this work, we utilize the \emph{Multi-Armed Bandits} paradigm to select the best intent or a set of intents for the current user. We assume that each intent cluster is an arm in the multi-armed bandit setting, and we aim to estimate the arm that returns the best overall candidates (i.e. highest expected \emph{quality score})\footnote{~ While the arms, in the general case, are to be independent from each other, we ignore this constraint within our work at this point. Obviously, the arms in our case are intent clusters which have overlapping candidates, however, as explained in \S~\ref{subsubsec:utilizing_intent_clusters}, we strive to increase the number of distinct elements in the earlier ranks. Furthermore, once the terms for the arms are selected via topic modeling, the ranking within, and serving a candidate from, each intent cluster is independent of one another.}. Multi-Armed Bandits are utilized commonly to deal with the \emph{explore-exploit} dilemma, and the framework is inspired by a setting where we have a number of slot machines (arms) in a casino which can be pulled one-by-one at each time step. The solution deals with how to (how many times, and in which order) play these arms in order to get the best rewards. In the traditional version of the problem, the user initially does not know which arm is the most beneficial at each step. If we assume that each arm returns its rewards from a static distribution, the user needs a certain time (via trials and errors) to learn (\emph{explore}) these distributions. Once the distributions for arms are learned, then it is optimal for the user to always play the arm with the highest mean reward (\emph{exploit}), to get the highest  utility. While the ultimate objective for a multi-armed bandit setting is to maximize the total rewards over a period of time via selecting the most beneficial arms, an arm selection strategy is often evaluated by \emph{regret} metric. Regret is defined as the expected utility difference of a specific arm selection strategy versus the strategy which always picks the best arm, and can be empirically calculated for a policy $p$ as:
\[
\textrm{regret(p)} = \frac{\sum_{t=1}^{T} r^*_t - r^p_t}{T} ~,
\]
where $r^*_t$ is the reward returned due to the decision of the optimal policy (i.e. the reward returned by the arm selected by the optimal policy) at time $t$, and $r^p_t$ is the reward returned due to the decision of policy $p$. In terms of expected rewards, the above formulation is equivalent to $\mathbb{E}[r^*] - \mathbb{E}[r^p]$.

Naturally, in selecting the set of best intent clusters (arms) to recommend candidates from (since candidates are recommended one-by-one, we choose a next intent/arm at each recommendation step, and get feedback from the user on the candidate served by the chosen arm), we also aim to minimize the regret, i.e. we want to pinpoint the most appropriate intent clusters as soon as possible within the session so that we can provide the most relevant results earlier in the ranks (this helps in improving Eq.~\ref{eq:p_opt} also, which is our main goal). To solve the regret minimization problem for multi-armed bandits, many algorithms have been proposed, where the most commonly used ones are based on \emph{Upper Confidence Bound} (UCB) \cite{auer_2002} on the mean rewards of arms (e.g. UCB1) and \emph{Thompson Sampling} \cite{thompson_1933, agrawal_2012}. While we do not provide the details of these algorithms here (see footnote\footnote{~ In simplified terms, UCB1 estimates a confidence interval (CI) on the expected rewards that would be received from an arm. If an arm has been pulled a small number of times, the CI would be wide, therefore the upper end (confidence bound, hence UCB) of this interval would be large. Since the algorithm chooses the next arm as the one with highest upper confidence bound, it motivates for exploration of less utilized arms. Thompson Sampling, on the other hand, aims to choose the next arm according to the probability that it has the highest expected reward among the other arms. Each time a reward is observed after the pull of an arm, the algorithm updates its belief on the mean rewards distribution of the arms.} for a simplified intuition), a performance comparison of the methodologies is presented in \S~\ref{sec:results}. Like UCB1 and Thompson Sampling, most of the arm selection policies assume a reward in the interval [0.0, 1.0]. This matches well with our application, since we can take the feedback as $0.0$ for \emph{Not Good}, and $1.0$ for the cases when the user gives the feedback as \emph{Good} for the recommended candidate.

\subsubsection{Variations of Multi-Armed Bandits Setting and Relation to Current Work}
There have been multiple variations on the multi-armed bandits setting that are application specific, most notably \emph{mortal bandits} \cite{chakrabarti_2008} (where each arm has a life-span, i.e. birth and death), and contextual bandits \cite{wang_2005, langford_2008} (where the arm rewards are dependent on the current context).

Our use case of multi-armed bandits align most closely with the contextual bandit algorithms. We utilize offline data to come up with a segmentation of the candidate space individually for each position, therefore we do utilize the position information as the context for multi-armed bandits setting. Furthermore, our ranking model takes user features \cite{hathuc2015expertisesearch} into account as well (via Eq.~\ref{eq:convex_score}), hence also utilizes the user context. The main difference of our work compared to the contextual bandit framework is that the context (user's intent of what kind of candidates s/he is looking for) remains the same within a session.

\subsection{Online Update of Intent Clusters}
\label{subsec:online_learning_term_weights}
One final effort we apply within our work is the improvement of intent clusters via utilizing the user feedback. As shown in Eq.~\ref{eq:linear_score}, we employ a linear scoring over the candidates returned by an intent cluster (i.e. \emph{matchScore}, utilized in Eq.~\ref{eq:convex_score}). Ideally, this formulation will give higher values for those candidates that the user would \emph{like} and low values for the others. Therefore, we should be able to update the intent cluster vector (i.e. $\vec{w}_{t_n}$ into $\vec{w'}_{t_n}$) after receiving feedback for each recommended candidate as follows:
\[
\vec{w'}_{t_n} = \vec{w}_{t_n} - \eta \cdot (\vec{w}_{t_n} \bullet \vec{w}_{c_m} - y_{c_m}) \cdot \vec{w}_{c_m} ~,
\]
where $\eta$ is the \emph{learning rate}, $y_{c_m}$ is the feedback of the user to the latest candidate ($c_m$) recommended from the intent cluster $t_n$, and similar to the notation of Eq.~\ref{eq:linear_score}, $\vec{w}_{c_m}$ is the binary vector of the candidate over the possible properties. This is the update methodology that would be used by the Stochastic Gradient Descent algorithm if we were solving a regression problem (i.e. $\vec{w}_{t_n} \bullet \vec{w}_{c_m}$ to estimate $y_{c_m}$) while optimizing mean squared error.

The main problem with the above update formulation is in the semantics of the optimization. The linear score is never meant to \emph{estimate} the user response, but rather to rank the candidates due to their similarity with the intent cluster. In the light of these points, we employ the following update formulation (which is similar to the \emph{perceptron} algorithm of Rosenblatt \cite{rosenblatt_1958}):
\begin{equation}
\label{eq:weights_update_perceptron}
\vec{w'}_{t_n} = \vec{w}_{t_n} + \eta \cdot y_{c_m} \cdot \vec{w}_{c_m} ~.
\end{equation}

In essence, the above update strategy aims to maximize:
\[
\sum_{c_i \mid y_{c_i} > 0} \vec{w}_{t_n} \cdot \vec{w}_{c_i} - \sum_{c_i \mid y_{c_i} \leq 0} \vec{w}_{t_n} \cdot \vec{w}_{c_i}
\]
over $w_{t_n}$ in an online manner, with the starting point (initial $w_{t_n}$, i.e. intent cluster vector) coming from the offline learned topic as a prior (the effect of the prior is diminished with large $\eta$). Therefore, it has the effect of getting the intent cluster as similar as possible (in terms of weights) to the positively rated candidates from that cluster (hence getting the candidates similar to the good ones higher in the ranks), while making it less and less similar (according to $\eta$) from those candidates that the user deemed not a good fit (moving them lower in the ranks of the intent cluster). This updating scheme also brings a new interpretation to Eq.~\ref{eq:convex_score}, and $\alpha$ within the equation. Basically, now, Eq.~\ref{eq:convex_score} presents a mixture (hence $\alpha$ being the \emph{mixture rate}) between an offline learned model (\emph{offlineScore}) and the online updated model (\emph{matchScore}), and $\alpha$ determines the amount of personalization we apply, since \emph{offlineScore} is the output of a model learned over all users, and \emph{matchScore} is updated within session, for a specific user (recruiter). Please note that $y_{c_m}$ should be 1.0 (good fit) or -1.0 (not a good fit) in the above formulation, and is not within range [0.0, 1.0] as was the case in \S~\ref{subsec:mab_clus}.

Indeed, such an update scheme is commonly used in the \emph{Online Learning} domain \cite{shalev_shwartz_2011, kevin_murphy_book_online_learning, romero_2013_online_learning}, and is aimed at getting the recommendation model closer to the current application's restrictions, which, in our case, are the user's preferences. We demonstrate the benefits of this update methodology in \S~\ref{sec:results} (along with the effect of modifying $\alpha$ and $\eta$).

\subsection{Overall Flow}
\label{subsec:methodology_summary}
\begin{figure}
\centering
\includegraphics[width=1.9in]{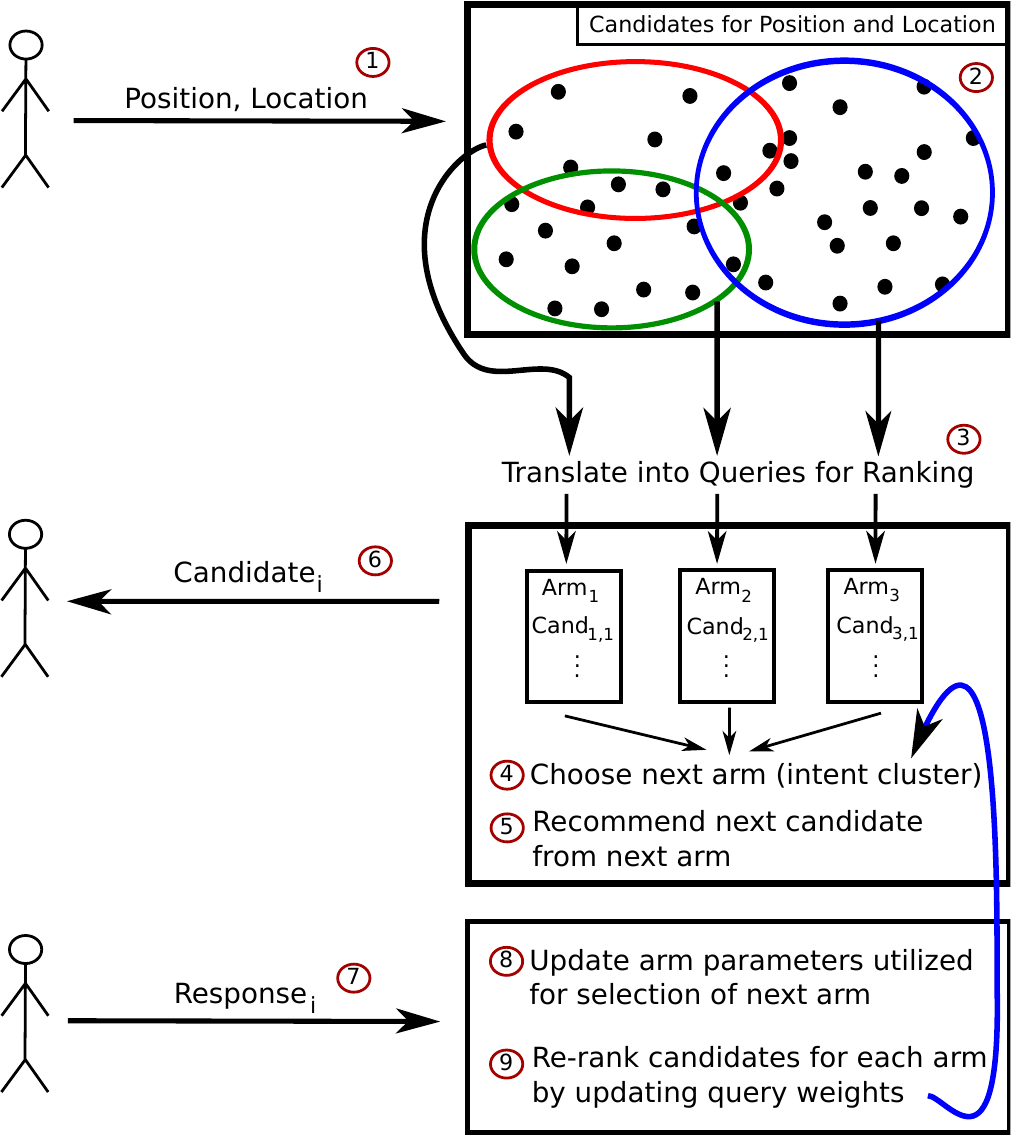}
\vspace{-10pt}
\caption{Recommendation System Flow. Numbers in the circles indicate the order of events in the flow.}
\label{fig:rec_workflow}
\end{figure}

Figure~\ref{fig:rec_workflow} presents an overall summary of our methodology as a flow of events in the recommendation process. Our workflow starts with the user providing an uninformative query, which is a position/title to be filled (step 1 in figure). Then, we reach into the candidate space that fits this title and segment the set of candidates into possibly overlapping \emph{intent clusters} (step 2), in our case via topic modeling (\S~\ref{subsubsec:lda_detailed}), which is performed offline (i.e. the topics/intent cluster properties for each title are predetermined and stored offline and segmentation is done online by picking up the stored intent clusters for user entered title search query). The intent clusters are translated into a set of queries (step 3), which are utilized to rank the candidates according to the term weights (\S~\ref{subsubsec:utilizing_intent_clusters}).

Selecting the most appropriate intent cluster (step 4), according to the current preferences of the user, is achieved via multi-armed bandits (\S~\ref{subsec:mab_clus}). With each newly shown candidate from the chosen intent cluster (steps 5 and 6), we can utilize the feedback (step 7) of the user on that specific candidate in order to both update the multi-armed bandit parameters (e.g. mean rewards) (step 8), and the term weights (step 9) so that we can get a better ranking for each intent cluster (\S~\ref{subsec:online_learning_term_weights}).

This overall recommendation scheme has been implemented within the LinkedIn Talent Services products, and we present the results of our online experiments in \S~\ref{sec:results}.

\section{Implementation Details}
\label{sec:implementation}

In this section, we focus on our offline pipeline which generates the intent clusters (\S~\ref{subsec:lda_clus}). The rest of our online framework has already been presented in \S~\ref{subsec:methodology_summary}.

\begin{figure} [htb]
\centering
\includegraphics[width=2.5in]{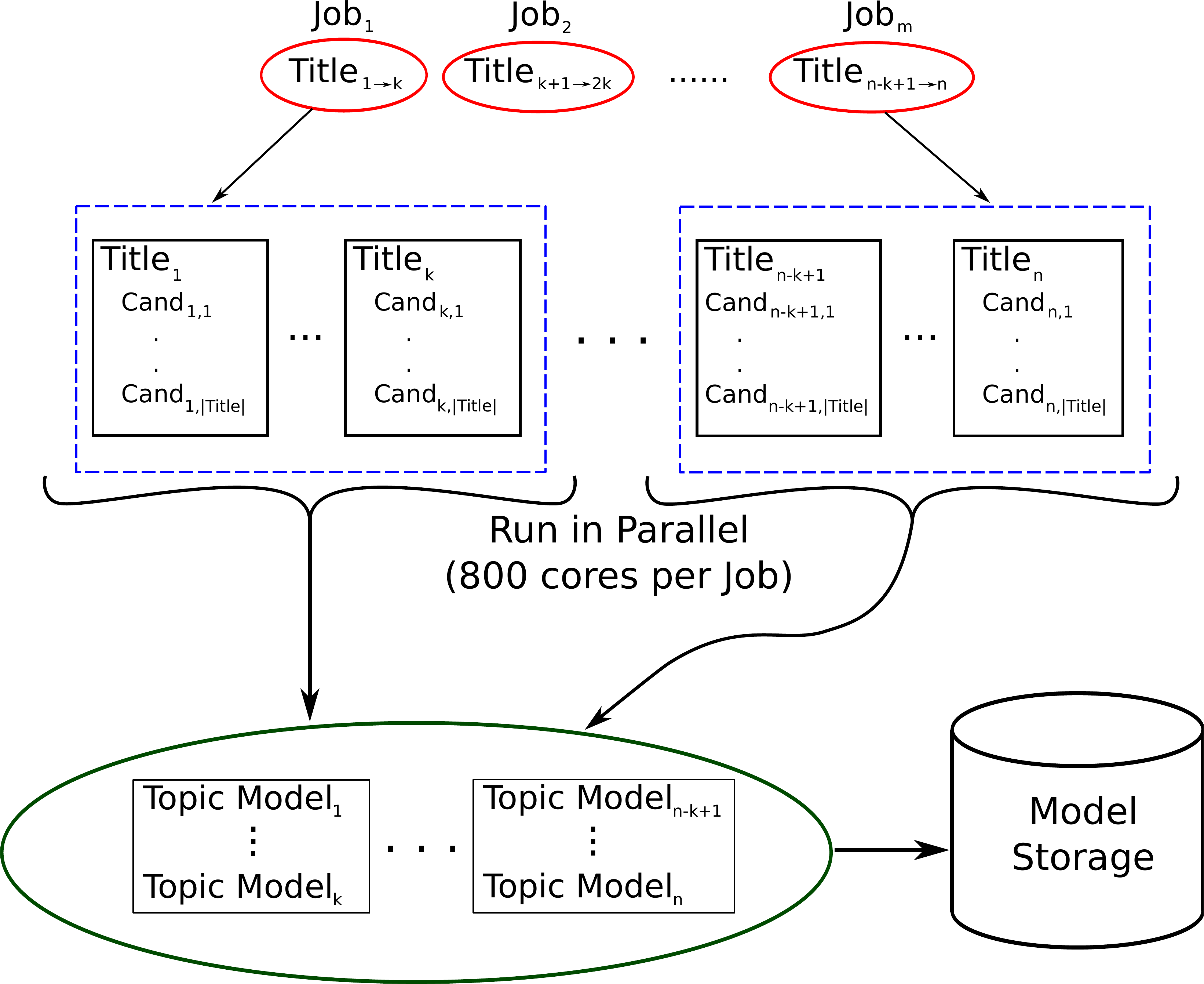}
\vspace{-10pt}
\caption{Architecture for Generating Intent Clusters}
\label{fig:offline_architecture}
\end{figure}

The main challenge with the application of LDA approach over each position is due to the scale we need to deal with. While there are methodologies \cite{li_2014, gao_2015} that aim to implement LDA to deal efficiently with large datasets, the main issue with our case is the tens of thousands of standardized titles we need to separately train a topic model for. The architecture that we employ for our offline intent cluster generation is given in Figure~\ref{fig:offline_architecture}. As it can be seen, we have a hierarchical separation of the titles into several jobs, and each job generates, in parallel, a topic model for each title, utilizing the properties of the candidates that belong to those titles. The reason for such a separation is two folds: (i) Sequential training of topic models is too costly due to the number of titles, (ii) Parallelizing all titles, each with a single job, brings too much overhead. Generated topic models are loaded into an online storage, to be served during run-time.

Currently, each run utilizes the immediate last six months period where we collect the candidates that started a new job with each specific title. This gives us around 1K-75K new candidates for each title. We run 10 jobs in parallel, where each job constructs a topic model for a subset of the titles, with around 800 cores per job. Our overall process takes around 17 hours to finish. Please note that the online selection and update of the clusters/topics are still in real-time. However, initial set of topics (utilized via MAB and updated online due to immediate feedback) are generated offline, according to the process outlined in this section.

\section{Results}
\label{sec:results}
In this section, we present both offline results of our proposed methodology, looking into the effect of changing the parameters of the system, as well as initial online results from the deployment of the scheme within our talent recommendation products.

\subsection{Offline Experiments} \label{sec:offl_exp}
\begin{figure}[!t]
\centering
\includegraphics[width=2.85in]{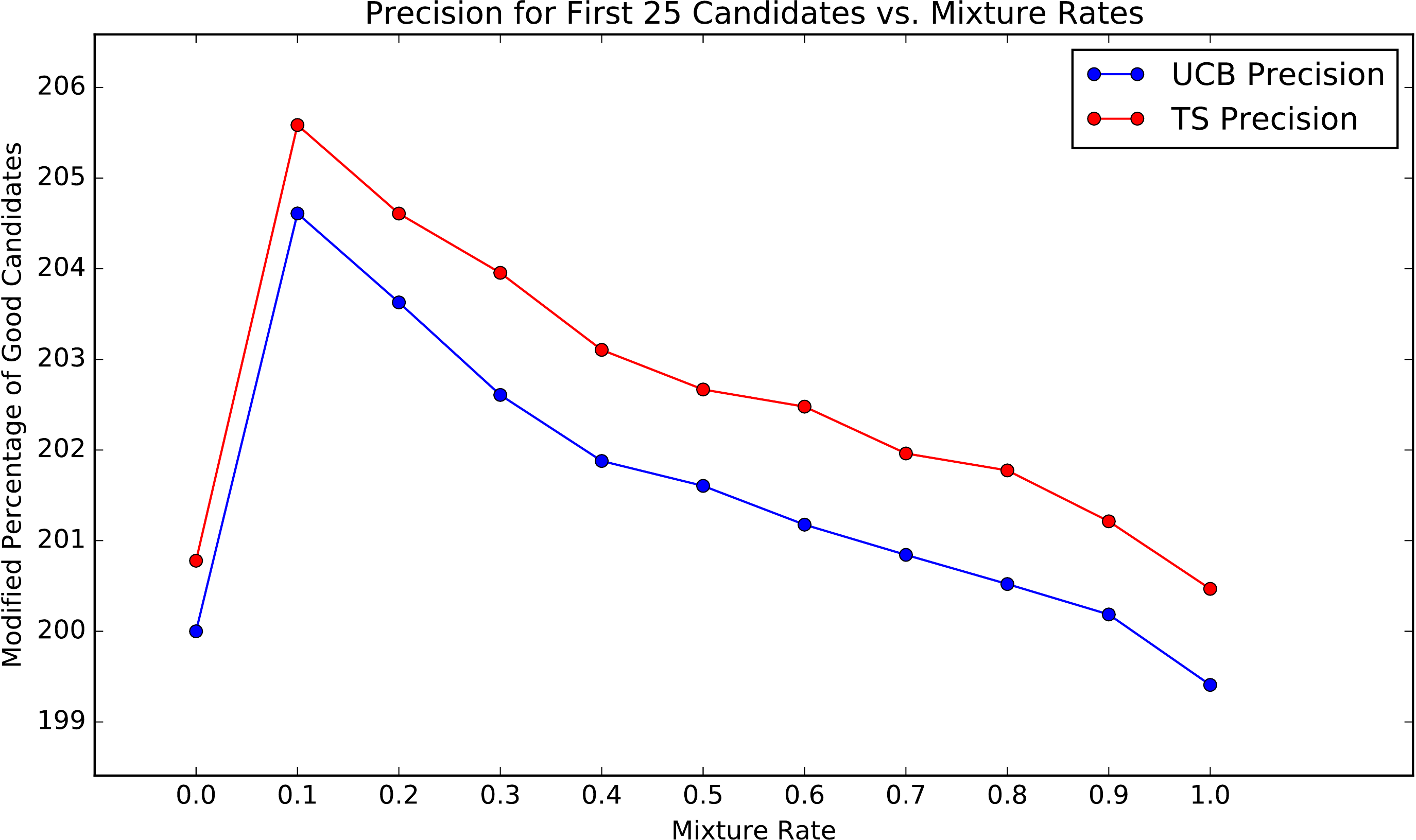}
\vspace{-10pt}
\caption{Precision@25 over Mixture Rate $\alpha$ (Averaged over All Explored Learning Rates)}
\label{fig:mrate}
\end{figure}

\begin{figure}[!t]
\centering
\includegraphics[width=2.85in]{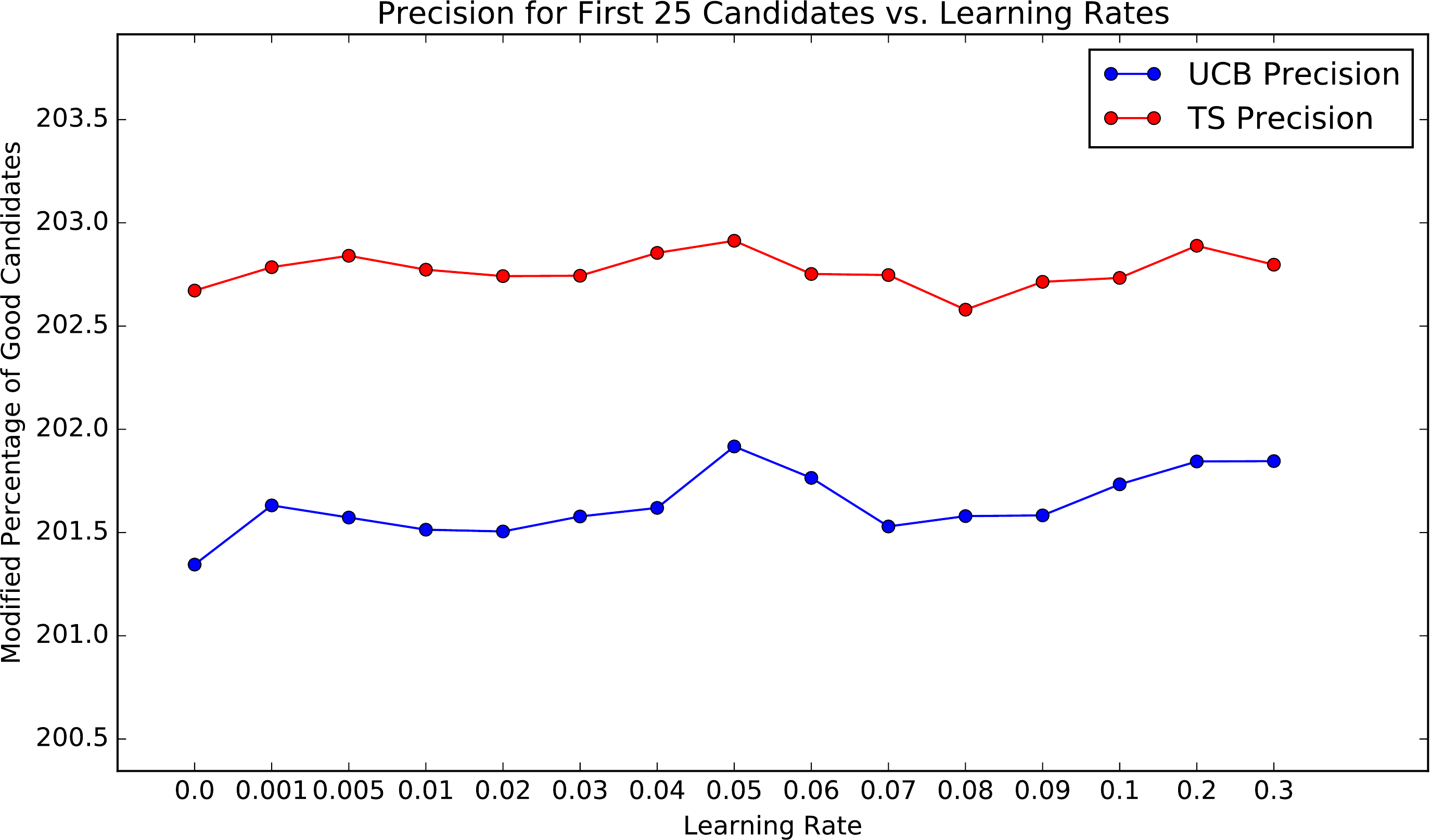}
\vspace{-10pt}
\caption{Precision@25 over Learning Rate $\eta$ (Averaged over All Explored Mixture Rates)}
\label{fig:lrate}
\end{figure}

\begin{figure}[!t]
\centering
\includegraphics[width=2.60in]{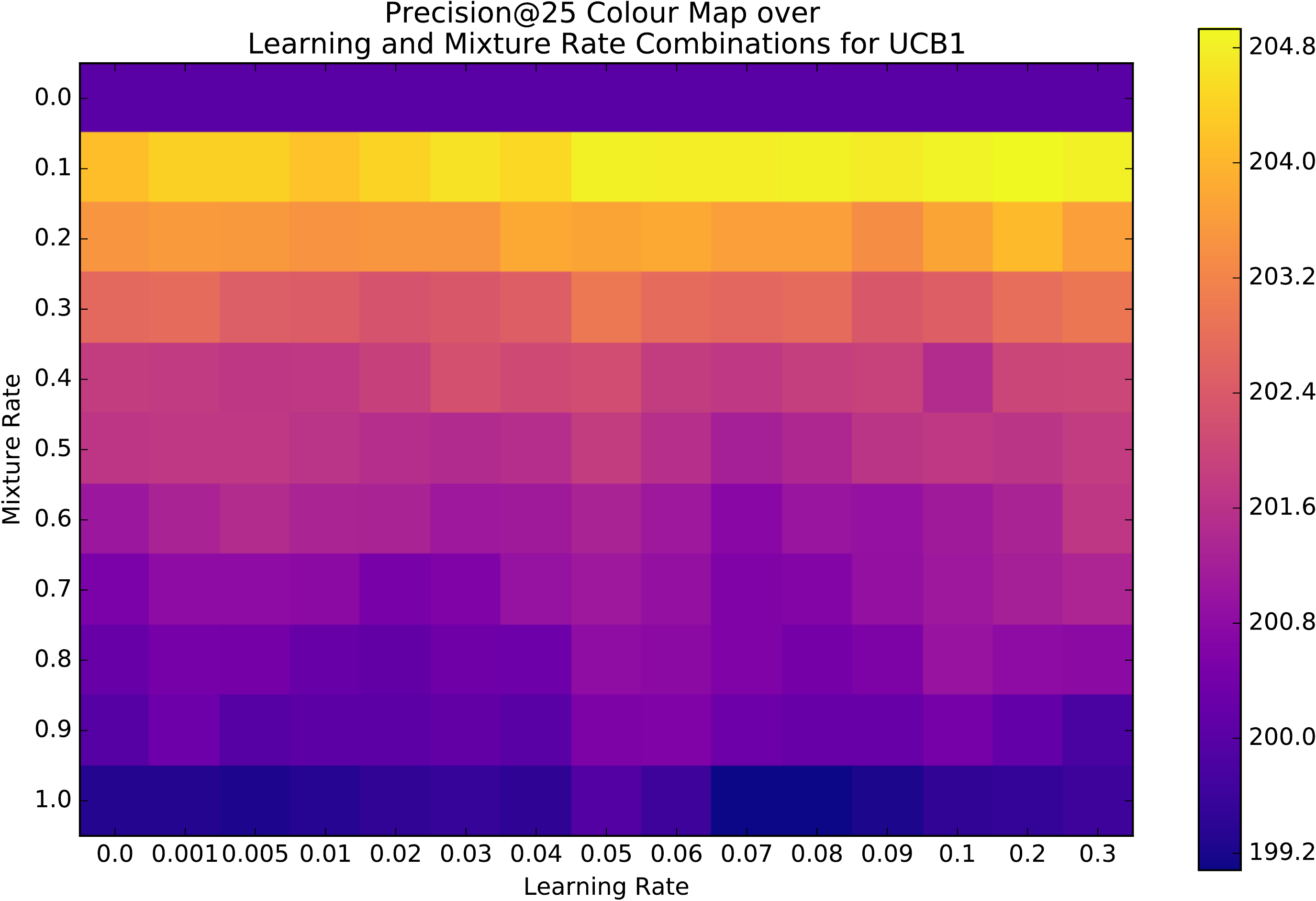}
\vspace{-10pt}
\caption{Precision@25 Colour Map over Learning and Mixture Rate Combinations for UCB1}
\label{fig:col_ucb}
\end{figure}

\begin{figure}[!t]
\centering
\includegraphics[width=2.60in]{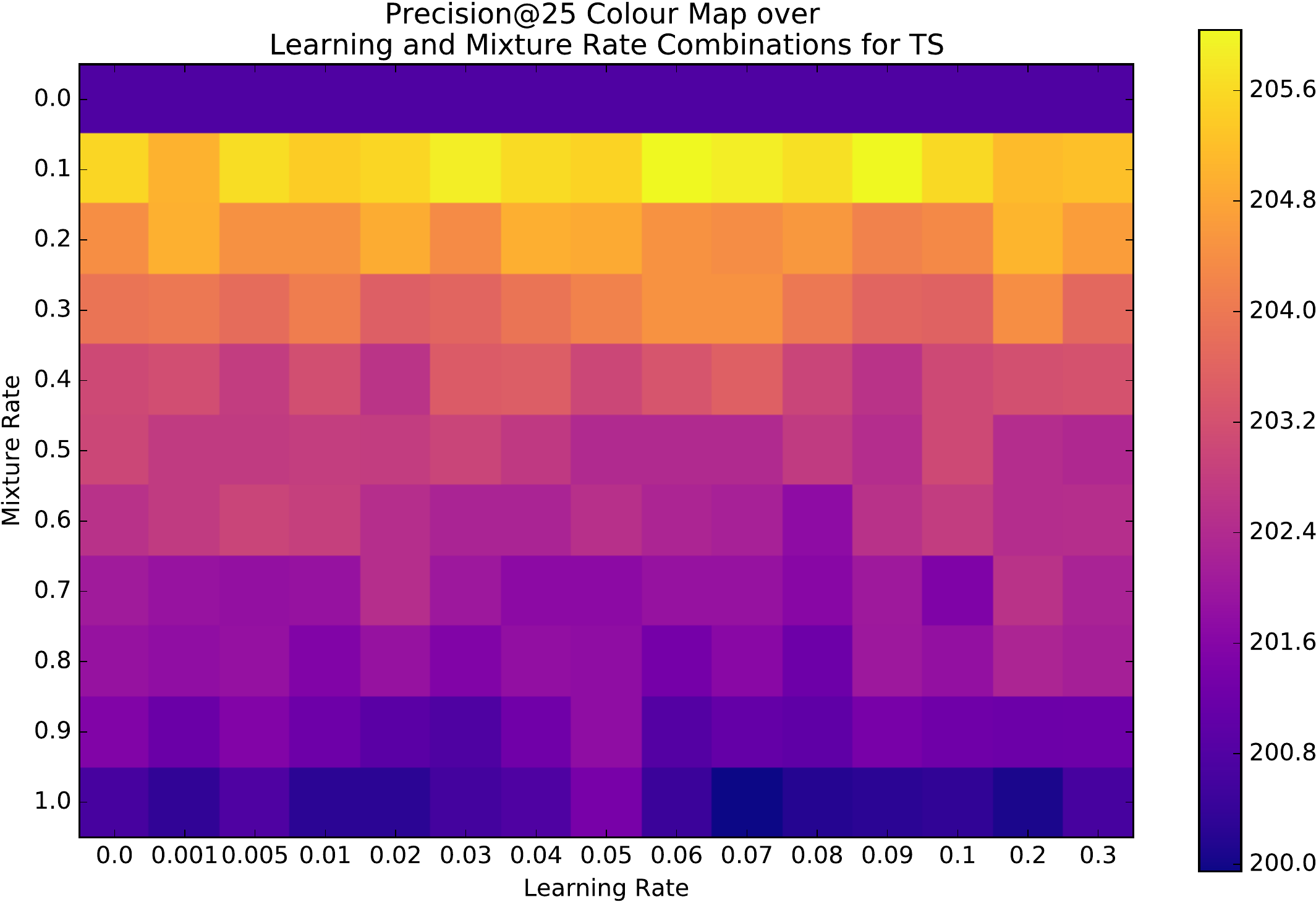}
\vspace{-10pt}
\caption{Precision@25 Colour Map over Learning and Mixture Rate Combinations for Thompson Sampling (TS)}
\label{fig:col_ts}
\end{figure}

We first evaluate our proposed methodology and the effect of mixture rate ($\alpha$ in Eq.~\ref{eq:convex_score}), learning rate ($\eta$ in Eq.~\ref{eq:weights_update_perceptron}), and the choice of MAB algorithm in an offline setting. For this purpose, we utilized the click logs from our \emph{Recruiter} application \cite{hathuc2015expertisesearch, hathuc2016talentsearch} over a period of 10 days within 2017. The dataset consists of a sampled set of user search instances along with a ranked list of candidates recommended and whether each candidate was clicked or not (search instances in the order of thousands were sampled for evaluation with hundreds of thousands of candidates recommended).  Each search instance is therefore a stream of candidates recommended to the recruiter for the same search query, which we re-rank offline using our proposed methodology and look at whether the methodology places the positively rated candidates higher. Figures \ref{fig:mrate} through \ref{fig:col_ts} presents the results of our offline evaluations.

In Figures \ref{fig:mrate} and \ref{fig:lrate}, we demonstrate the precision (percentage of positively rated candidates\footnote{~ We modify the precision values with a constant multiplier for company policy.}) over different mixture rate values and learning rates for both UCB1 and Thompson Sampling (TS) arm selection algorithm choices, focusing on the first 25 candidates recommended. Each point in the graphs represent an average over different parameters, e.g. the precision value for learning rate $\eta = 0.01$ in Figure~\ref{fig:mrate} is calculated by changing the mixture rate over a specific set of values, given in the x-axis of Figure~\ref{fig:lrate} to be exact, and getting the average precision over different runs. We can see from the figures that learning rate has a peaking behavior (at 0.05), and mixing the offline score with online updated \emph{matchScore} is necessary ($\alpha$=0 gives the worst results). However, further increasing the mixture rate ($\alpha$) reduces the precision, which indicates the need for balancing between the global offline learned model (\emph{offlineScore} in Eq.~\ref{eq:convex_score}) and personalized, online updated model (\emph{matchScore} in Eq.~\ref{eq:convex_score}). Also, Thompson Sampling performs better overall, compared to UCB1, therefore it has been our choice for online deployment. Finally, to allow for a more granular examination of our results, we provide a heat map of all paired combinations of learning and mixture rates in Figures \ref{fig:col_ucb} and \ref{fig:col_ts}.

\begin{figure}[!t]
\centering
\includegraphics[width=2.9in]{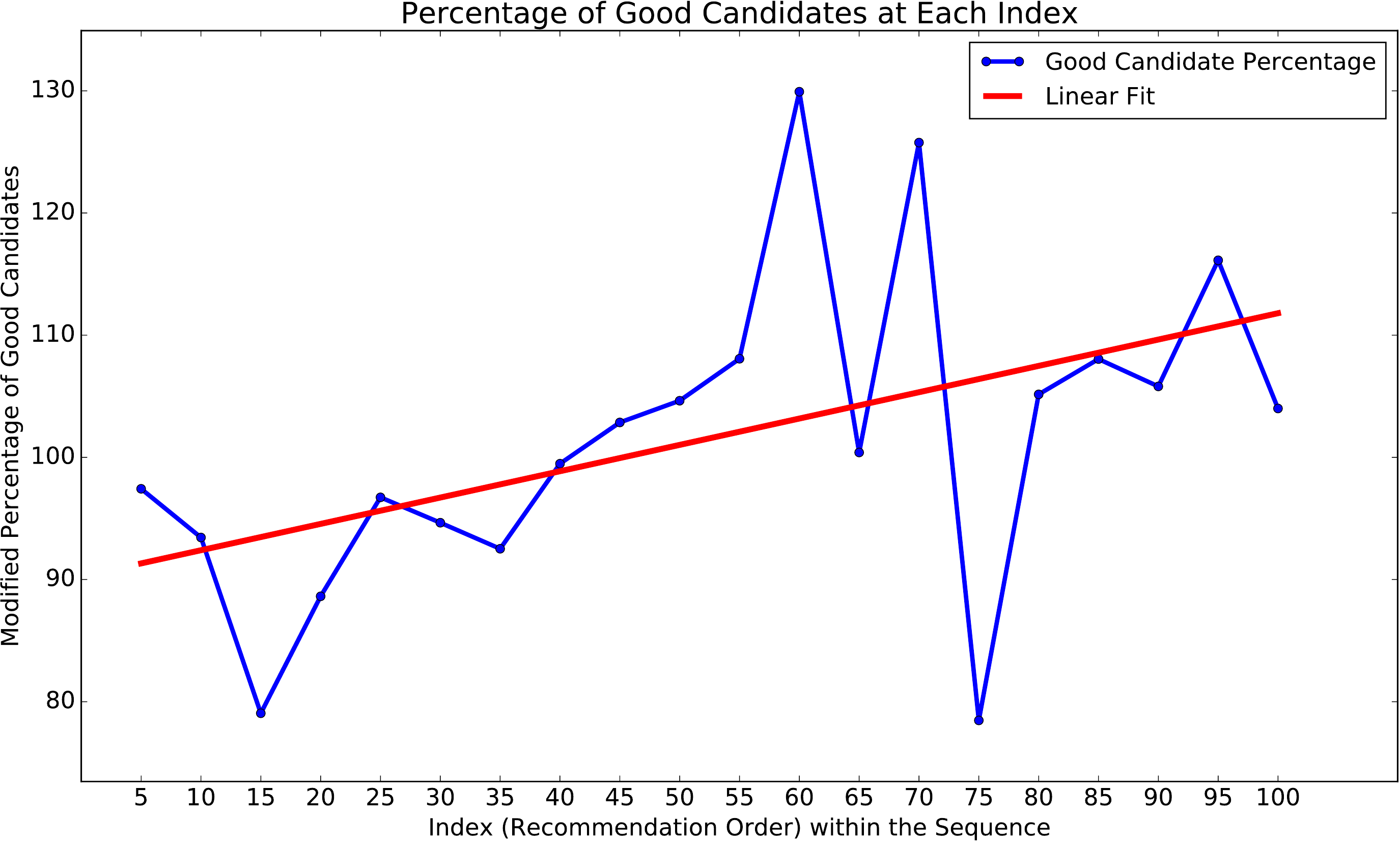}
\vspace{-10pt}
\caption{Evolution of Positive Feedback Percentage over the Life Cycle of Sessions, with the Trend-line in Red (x-axis gives the index within which we do the recommendation, i.e. 5 $\rightarrow$ candidates recommended in 1-5th ranks, 10 $\rightarrow$ 6-10th, 60 $\rightarrow$ 56-60th etc.).}
\label{fig:interested_perf_5}
\end{figure}

\subsection{Online Results}
\begin{figure}[!t]
\centering
\includegraphics[width=2.3in]{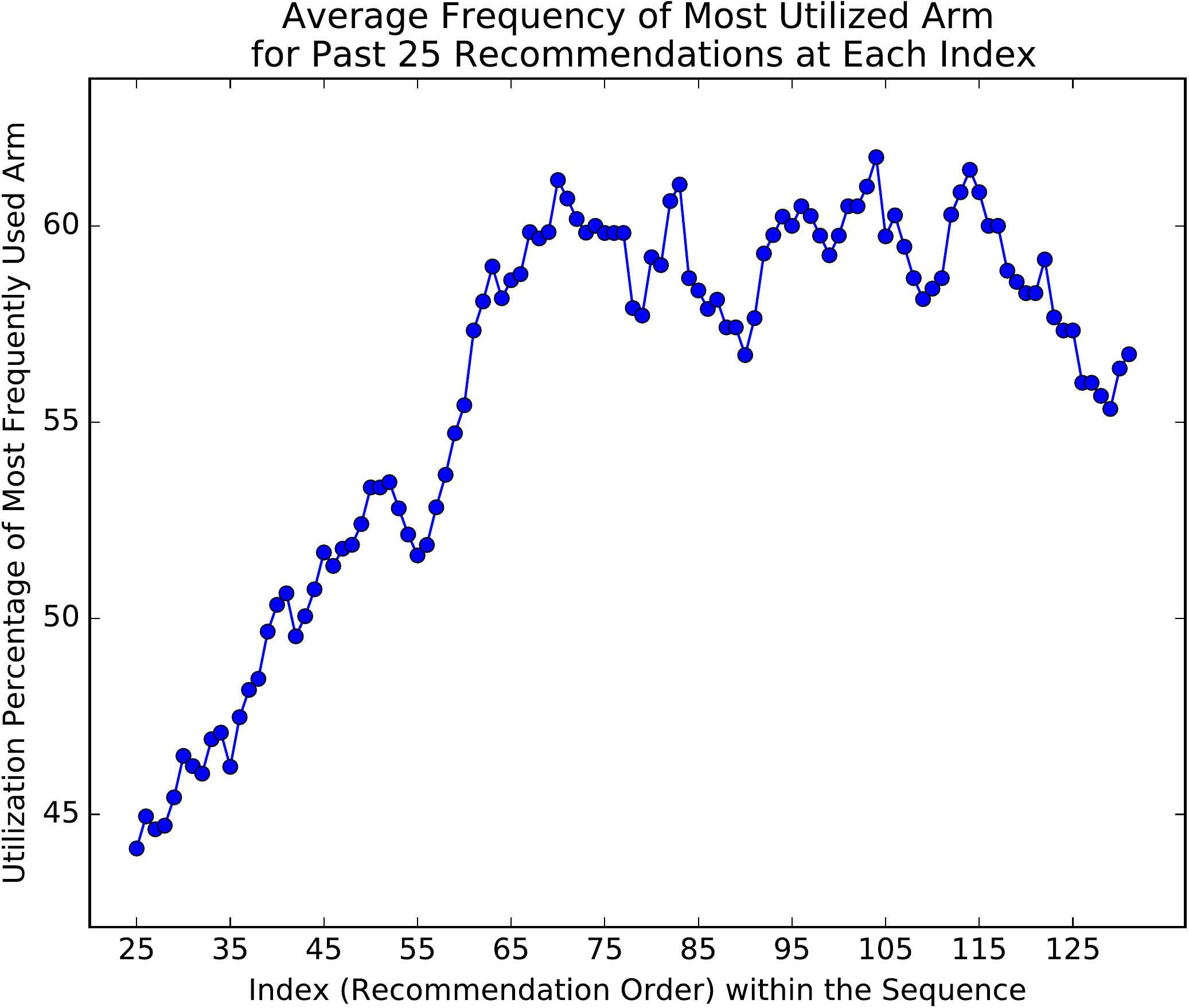}
\vspace{-10pt}
\caption{Percentage of Most Frequently Utilized Arm over the Life Cycle of Sessions}
\label{fig:frequency_arm_conv}
\end{figure}

\begin{figure}[!t]
\centering
\includegraphics[width=2.3in]{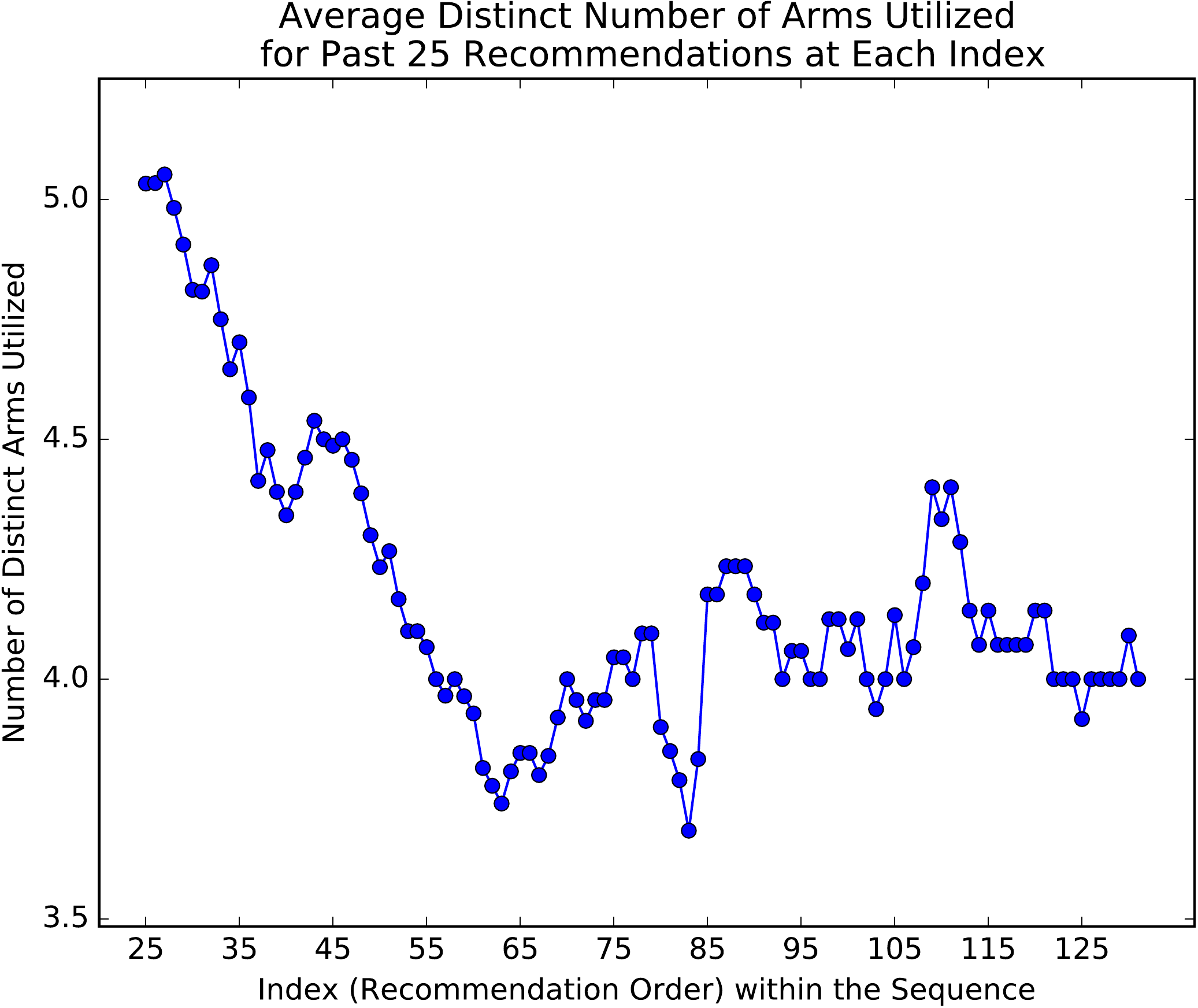}
\vspace{-10pt}
\caption{Number of Distinct Arms Utilized Arm over the Life Cycle of Sessions}
\label{fig:distinct_arm_conv}
\end{figure}

Next, we would like to present the results from our online deployment of the proposed methodology within 2017. The number of search sessions we have utilized for the evaluation is in the order of hundreds, where we received feedback for thousands of recommended candidates. The results are given in Figures \ref{fig:interested_perf_5} through \ref{fig:distinct_arm_conv}. In Figure~\ref{fig:interested_perf_5}, we present the precision results (modified similar to the results presented in \S~\ref{sec:offl_exp}), averaged over all users, during the evolution of the search session (due to online learning), where we also provide the trend-line. The improvement in precision as we get more feedback is visible from the graph, which demonstrates the online learning capabilities of the proposed scheme.

We also examined the convergence behavior of the multi-armed bandits utilized to select intent clusters, as each search session progresses. Figure~\ref{fig:frequency_arm_conv} shows that, as expected, the utilization (pull) percentage of the most frequently used (pulled) arm (which represents an intent cluster) increases as we get more and more feedback within a search session (we calculated these statistics within a sliding window of 25 at each newly recommended candidate, i.e. looking at the utilized arms for the past 25 candidates at each rank). Finally, Figure~\ref{fig:distinct_arm_conv} (where the statistics are calculated using a sliding window of 25, similar to Figure~\ref{fig:frequency_arm_conv}) also supports our previous observation, where it can be noticed that the number of unique arms utilized to recommend a candidate gets lower as we get more feedback within the session (which means that \emph{exploration} over intent queries lessens as we learn the user preferences, and \emph{exploitation} kicks in more and more as the session progresses).

\section{Related Work}
\label{sec:related_work}

\subsection{Topic Modeling in Recommender Systems}
To the best of our knowledge, our work is the first application of topic modeling within the domain of recruiting and candidate recommendation for hiring. Due to this fact, we will only explore a limited number of works on topic modeling for recommender systems within this section.

LDA and its variants are widely used in applications to extract topics from items and users in recommender systems, mainly for \emph{personalized document recommendation} \cite{Chang, Luostarinen}. Some sample work that diverges from this trend is given in \cite{LCARS} and \cite{Auralist}, which focus on social event recommendations and musical recommendations, respectively. The authors of \cite{LCARS} utilize LDA for event and venue recommendation, proposing a location and content-aware recommender system which gives consideration to both personal interest and local preference. Finally, \cite{Auralist} introduces \emph{Auralist} system which aims to inject serendipity, novelty and diversity into recommendations while limiting the impact on accuracy.

\subsection{Multi-Armed Bandits in \\Recommender Systems}
While there are several works which apply multi-armed bandits within recommender systems domain, we would like to do a deeper examination of the most relevant set of previous literature. An earlier work \cite{radlinski_2008} focuses on the need for diversification of recommended documents, since the relevance scores of ranked items are not independent of each other. Their contribution is \emph{ranked bandits} algorithm, which assigns separate multi-armed bandits for each ranked object in the recommended sequence. This is strictly different compared to our methodology, since we aim to find the best ranking for the current user, and our arms are ranking algorithms, not documents. Therefore choosing the next arm is equivalent in our case to choosing the remaining best item (i.e. next rank) from the arm. A successor paper is given in \cite{kohli_2013}, where the difference is in the reward function. \cite{radlinski_2008} assumes a positive reward for only the first relevant document, where \cite{kohli_2013} receives positive rewards for all clicked documents in the generated ranking, which is more suitable to our application.

The last work we would like to mention here is given in \cite{hofmann_2011} where the authors aim to apply a contextual bandit setting for recommending articles, where the utility of an arm is based on a feature vector that describes the query and the user properties. They also argue that a fully exploitative approach would only collect information of the top ranked documents, hence they propose an approach of interleaved lists where the recommendation shown to user comes from both an explorative and an exploitative list. This is similar to our proposed approach, however we utilize each ranking per session for both exploration and exploitation.

\subsection{Online Session Personalization in Recommender Systems}
Historically, content neutral collaborative filtering and hybrid collaborative filtering systems have been popular with online services such as Amazon, eBay, and Spotify, and continue to be used for many applications. Although early collaborative filtering systems used memory-based methods based on item-item similarity metrics such as cosine similarity, most modern collaborative filtering systems use matrix factorization algorithms. The classical matrix factorization methods for collaborative filtering include Singular Value Decomposition (SVD), Principal Component Analysis (PCA), and Probabilistic Matrix Factorization (PMF). All of these methods suffer from the limitation that online gradient descent is impractical for large matrices, and most of these methods have been forced to respond to user actions through re-training.

Factorization machines \cite{rendle2010fm} are a popular new approach to personalized recommendation systems in the context of sparse data which generalize previous factorization methods for collaborative filtering. Although the initial publications on factorization machines addressed the problem solely in the context of offline learning, later work \cite{kitazawa2016, lu2013second_order_filtering} addresses methods of online learning which are applicable to factorization machines. 

In \cite{gemmell2008folksonomies}, the authors present a different sort of online personalization system for free-form tag based collaborative filtering systems in which tags are assembled into clusters and the user's interests are modeled as a vector of cluster weights. The authors define the weight of each cluster as the ratio of the number of times a user has annotated a resource described by one or more tags in a cluster to the total number of annotations a user has made. As the user interacts with resources the weights can be recomputed trivially. The similarity with our work comes from the way the authors model a user's interaction with a large space of labels by clustering the labels and assigning a dynamically computed weight to the user's affinity for the entire cluster. However, their method of personalization is significantly different from our approach since we allow for both positive and negative feedback, and update our vector representation of the user's interests (via intent cluster updates) rather than merely rely on the number of times a user interacts with a resource from a cluster.

Linear models are also widely used in many recommendation systems when content aware features are desirable, but models trained on large enough sample sizes of user-item interactions to estimate the likelihood of users interacting with resources can lack personalization. Generalized linear mixed models (GLMix) can be used to combine a traditional globally trained linear model with individually trained regression coefficients \cite{zhang2016glmix}. Online personalization is achieved by retraining the model frequently to capture recent user-resource interactions.

\section{Conclusions and Future Work}
\label{sec:conclusions}

In this work, we have introduced our model for recommending candidates to a user, where we utilize the user feedback to improve the candidate quality in real-time and in-session. Such a system allows for cases where the user provided information is limited to only a position to be filled. We have presented our methodology which utilizes meaningful segments (intent clusters) of the candidate space, combined with multi-armed bandits for segment selection and online update of the segments in-session due to user feedback on candidates which are recommended one at a time. We presented offline results on real-world recruiting data, as well as online results for our deployment of the proposed approach, which demonstrate the advantages of our approach, and the effect of online learning within the domain of talent search.

The potential areas of future work are improving the convergence speed to a specific user intent, and increasing the diversity of recommended candidates. For improving convergence, we plan to investigate warm-start models which utilize the behavior of the similar users to give a prior on the intent clusters (arms) before the recommendation session begins. Similarly, while within this work we only utilized explicit user rating feedback (good fit and bad fit, akin to thumbs up and thumbs down), distinguishing between different types of feedback would also be beneficial to relevant candidate recommendations to the user. Finally, we plan to explore generation of the intent clusters via hierarchical topic models \cite{griffiths_2004} to recommend a potentially more diverse set of candidates to the user.

\bibliographystyle{ACM-Reference-Format}
\bibliography{sigproc}


\begin{thebibliography}{35}


\ifx \showCODEN    \undefined \def \showCODEN     #1{\unskip}     \fi
\ifx \showDOI      \undefined \def \showDOI       #1{#1}\fi
\ifx \showISBNx    \undefined \def \showISBNx     #1{\unskip}     \fi
\ifx \showISBNxiii \undefined \def \showISBNxiii  #1{\unskip}     \fi
\ifx \showISSN     \undefined \def \showISSN      #1{\unskip}     \fi
\ifx \showLCCN     \undefined \def \showLCCN      #1{\unskip}     \fi
\ifx \shownote     \undefined \def \shownote      #1{#1}          \fi
\ifx \showarticletitle \undefined \def \showarticletitle #1{#1}   \fi
\ifx \showURL      \undefined \def \showURL       {\relax}        \fi
\providecommand\bibfield[2]{#2}
\providecommand\bibinfo[2]{#2}
\providecommand\natexlab[1]{#1}
\providecommand\showeprint[2][]{arXiv:#2}

\bibitem[\protect\citeauthoryear{Agrawal and Goyal}{Agrawal and Goyal}{2012}]%
        {agrawal_2012}
\bibfield{author}{\bibinfo{person}{S. Agrawal} {and} \bibinfo{person}{N.
  Goyal}.} \bibinfo{year}{2012}\natexlab{}.
\newblock \showarticletitle{Analysis of Thompson Sampling for the Multi-armed
  Bandit Problem}.
\newblock \bibinfo{journal}{\emph{JMLR}}  \bibinfo{volume}{23}
  (\bibinfo{year}{2012}).
\newblock


\bibitem[\protect\citeauthoryear{Aha}{Aha}{1992}]%
        {aha_1992}
\bibfield{author}{\bibinfo{person}{D. Aha}.} \bibinfo{year}{1992}\natexlab{}.
\newblock \showarticletitle{Generalizing from case studies: A case study}. In
  \bibinfo{booktitle}{\emph{ICML}}.
\newblock


\bibitem[\protect\citeauthoryear{Akaike}{Akaike}{1974}]%
        {akaike_1974}
\bibfield{author}{\bibinfo{person}{H. Akaike}.}
  \bibinfo{year}{1974}\natexlab{}.
\newblock \showarticletitle{A new look at the statistical model
  identification}.
\newblock \bibinfo{journal}{\emph{IEEE Trans. Automatic Control}}
  \bibinfo{volume}{19}, \bibinfo{number}{6} (\bibinfo{year}{1974}),
  \bibinfo{pages}{716--723}.
\newblock


\bibitem[\protect\citeauthoryear{Auer, Cesa-Bianchi, and Fischer}{Auer
  et~al\mbox{.}}{2002}]%
        {auer_2002}
\bibfield{author}{\bibinfo{person}{P. Auer}, \bibinfo{person}{N. Cesa-Bianchi},
  {and} \bibinfo{person}{P. Fischer}.} \bibinfo{year}{2002}\natexlab{}.
\newblock \showarticletitle{Finite-time Analysis of the Multiarmed Bandit
  Problem}.
\newblock \bibinfo{journal}{\emph{Machine Learning Journal}}
  \bibinfo{volume}{47} (\bibinfo{year}{2002}), \bibinfo{pages}{235--256}.
\newblock


\bibitem[\protect\citeauthoryear{Blei, Ng, and Jordan}{Blei
  et~al\mbox{.}}{2003}]%
        {blei_2003_lda}
\bibfield{author}{\bibinfo{person}{David~M. Blei}, \bibinfo{person}{Andrew~Y.
  Ng}, {and} \bibinfo{person}{Michael~I. Jordan}.}
  \bibinfo{year}{2003}\natexlab{}.
\newblock \showarticletitle{Latent Dirichlet Allocation}.
\newblock \bibinfo{journal}{\emph{JMLR}}  \bibinfo{volume}{3}
  (\bibinfo{year}{2003}), \bibinfo{pages}{993--1022}.
\newblock


\bibitem[\protect\citeauthoryear{Chakrabarti, Kumar, Radlinski, and
  Upfal}{Chakrabarti et~al\mbox{.}}{2008}]%
        {chakrabarti_2008}
\bibfield{author}{\bibinfo{person}{D. Chakrabarti}, \bibinfo{person}{R. Kumar},
  \bibinfo{person}{F. Radlinski}, {and} \bibinfo{person}{E. Upfal}.}
  \bibinfo{year}{2008}\natexlab{}.
\newblock \showarticletitle{Mortal multi-armed bandits}. In
  \bibinfo{booktitle}{\emph{NIPS}}. \bibinfo{pages}{273--280}.
\newblock


\bibitem[\protect\citeauthoryear{Chang and Feng}{Chang and Feng}{2013}]%
        {Chang}
\bibfield{author}{\bibinfo{person}{T.~M. Chang} {and} \bibinfo{person}{W.
  Feng}.} \bibinfo{year}{2013}\natexlab{}.
\newblock \showarticletitle{LDA-based Personalized Document Recommendation}. In
  \bibinfo{booktitle}{\emph{Proc. of PACIS 2013}}.
\newblock


\bibitem[\protect\citeauthoryear{Fontenla-Romero, Guijarro-Berdinas,
  Martinez-Rego, Perez-Sanchez, and Peteiro-Barral}{Fontenla-Romero
  et~al\mbox{.}}{2013}]%
        {romero_2013_online_learning}
\bibfield{author}{\bibinfo{person}{O. Fontenla-Romero}, \bibinfo{person}{B.
  Guijarro-Berdinas}, \bibinfo{person}{D. Martinez-Rego}, \bibinfo{person}{B.
  Perez-Sanchez}, {and} \bibinfo{person}{D. Peteiro-Barral}.}
  \bibinfo{year}{2013}\natexlab{}.
\newblock \showarticletitle{Online Machine Learning}.
\newblock \bibinfo{journal}{\emph{Efficiency and Scalability Methods for
  Computational Intellect}} (\bibinfo{year}{2013}), \bibinfo{pages}{27--54}.
\newblock


\bibitem[\protect\citeauthoryear{Gao, Sun, Wang, Liu, Yan, and Zeng}{Gao
  et~al\mbox{.}}{2015}]%
        {gao_2015}
\bibfield{author}{\bibinfo{person}{Yang Gao}, \bibinfo{person}{Zhenlong Sun},
  \bibinfo{person}{Yi Wang}, \bibinfo{person}{Xiaosheng Liu},
  \bibinfo{person}{Jianfeng Yan}, {and} \bibinfo{person}{Jia Zeng}.}
  \bibinfo{year}{2015}\natexlab{}.
\newblock \showarticletitle{A comparative study on parallel lda algorithms in
  mapreduce framework}.
\newblock \bibinfo{journal}{\emph{LNCS}}  \bibinfo{volume}{9078}
  (\bibinfo{year}{2015}), \bibinfo{pages}{675--689}.
\newblock


\bibitem[\protect\citeauthoryear{Gemmell, Shepitsen, Mobasher, and
  Burke}{Gemmell et~al\mbox{.}}{2008}]%
        {gemmell2008folksonomies}
\bibfield{author}{\bibinfo{person}{J. Gemmell}, \bibinfo{person}{A. Shepitsen},
  \bibinfo{person}{M. Mobasher}, {and} \bibinfo{person}{R. Burke}.}
  \bibinfo{year}{2008}\natexlab{}.
\newblock \showarticletitle{Personalization in Folksonomies Based on Tag
  Clustering}. In \bibinfo{booktitle}{\emph{Proc. of the 6th Workshop on
  Intelligent Techniques for Web Personalization and Recommender Systems}}.
\newblock


\bibitem[\protect\citeauthoryear{Griffiths, Jordan, and Tenenbaum}{Griffiths
  et~al\mbox{.}}{2004}]%
        {griffiths_2004}
\bibfield{author}{\bibinfo{person}{T. Griffiths}, \bibinfo{person}{M. Jordan},
  {and} \bibinfo{person}{J. Tenenbaum}.} \bibinfo{year}{2004}\natexlab{}.
\newblock \showarticletitle{Hierarchical Topic Models and the Nested Chinese
  Restaurant Process}. In \bibinfo{booktitle}{\emph{NIPS}}.
\newblock


\bibitem[\protect\citeauthoryear{Ha-Thuc, Venkataraman, Rodriguez, Sinha,
  Sundaram, and Guo}{Ha-Thuc et~al\mbox{.}}{2015}]%
        {hathuc2015expertisesearch}
\bibfield{author}{\bibinfo{person}{Viet Ha-Thuc}, \bibinfo{person}{Ganesh
  Venkataraman}, \bibinfo{person}{Mario Rodriguez}, \bibinfo{person}{Shakti
  Sinha}, \bibinfo{person}{Senthil Sundaram}, {and} \bibinfo{person}{Lin Guo}.}
  \bibinfo{year}{2015}\natexlab{}.
\newblock \showarticletitle{Personalized Expertise Search at LinkedIn}. In
  \bibinfo{booktitle}{\emph{Proc. of the 4th IEEE Int. Conf. Big Data}}.
\newblock


\bibitem[\protect\citeauthoryear{Ha-Thuc, Xu, Kanduri, Wu, Dialani, Yan, Gupta,
  and Sinha}{Ha-Thuc et~al\mbox{.}}{2016}]%
        {hathuc2016talentsearch}
\bibfield{author}{\bibinfo{person}{Viet Ha-Thuc}, \bibinfo{person}{Ye Xu},
  \bibinfo{person}{Satya~Pradeep Kanduri}, \bibinfo{person}{Xianren Wu},
  \bibinfo{person}{Vijay Dialani}, \bibinfo{person}{Yan Yan},
  \bibinfo{person}{Abhishek Gupta}, {and} \bibinfo{person}{Shakti Sinha}.}
  \bibinfo{year}{2016}\natexlab{}.
\newblock \showarticletitle{Search by Ideal Candidates: Next Generation of
  Talent Search at LinkedIn}. In \bibinfo{booktitle}{\emph{ACM WWW}}.
\newblock


\bibitem[\protect\citeauthoryear{Hofmann, Whiteson, and de~Rijke}{Hofmann
  et~al\mbox{.}}{2011}]%
        {hofmann_2011}
\bibfield{author}{\bibinfo{person}{K. Hofmann}, \bibinfo{person}{S. Whiteson},
  {and} \bibinfo{person}{M. de Rijke}.} \bibinfo{year}{2011}\natexlab{}.
\newblock \showarticletitle{Contextual Bandits for Information Retrieval}. In
  \bibinfo{booktitle}{\emph{NIPS}}.
\newblock


\bibitem[\protect\citeauthoryear{Jarvelin and Kekalainen}{Jarvelin and
  Kekalainen}{2002}]%
        {jarvelin_2002}
\bibfield{author}{\bibinfo{person}{K. Jarvelin} {and} \bibinfo{person}{J.
  Kekalainen}.} \bibinfo{year}{2002}\natexlab{}.
\newblock \showarticletitle{Cumulated Gain-based Evaluation of IR Techniques}.
\newblock \bibinfo{journal}{\emph{ACM Trans. on Information Systems (TOIS)}}
  \bibinfo{volume}{20}, \bibinfo{number}{4} (\bibinfo{year}{2002}),
  \bibinfo{pages}{422--446}.
\newblock


\bibitem[\protect\citeauthoryear{Jordan}{Jordan}{1999}]%
        {Jordan1999}
\bibfield{author}{\bibinfo{person}{M. Jordan}.}
  \bibinfo{year}{1999}\natexlab{}.
\newblock In \bibinfo{booktitle}{\emph{Learning in Graphical Models}}.
  \bibinfo{publisher}{MIT Press}.
\newblock


\bibitem[\protect\citeauthoryear{Kitazawa}{Kitazawa}{2016}]%
        {kitazawa2016}
\bibfield{author}{\bibinfo{person}{Takuya Kitazawa}.}
  \bibinfo{year}{2016}\natexlab{}.
\newblock \showarticletitle{Incremental Factorization Machines for Persistently
  Cold-starting Online Item Recommendation}.
\newblock \bibinfo{journal}{\emph{CoRR}}  \bibinfo{volume}{abs/1607.02858}
  (\bibinfo{year}{2016}).
\newblock
\urldef\tempurl%
\url{http://arxiv.org/abs/1607.02858}
\showURL{%
\tempurl}


\bibitem[\protect\citeauthoryear{Kohli, Salek, and Stoddard}{Kohli
  et~al\mbox{.}}{2013}]%
        {kohli_2013}
\bibfield{author}{\bibinfo{person}{P. Kohli}, \bibinfo{person}{M. Salek}, {and}
  \bibinfo{person}{G. Stoddard}.} \bibinfo{year}{2013}\natexlab{}.
\newblock \showarticletitle{A Fast Bandit Algorithm for Recommendations to
  Users with Heterogeneous Tastes}. In \bibinfo{booktitle}{\emph{AAAI}}.
\newblock


\bibitem[\protect\citeauthoryear{Kotthoff}{Kotthoff}{2014}]%
        {kotthoff_2014}
\bibfield{author}{\bibinfo{person}{L. Kotthoff}.}
  \bibinfo{year}{2014}\natexlab{}.
\newblock \showarticletitle{Algorithm selection for combinatorial search
  problems: A survey}.
\newblock \bibinfo{journal}{\emph{AI Magazine}} (\bibinfo{year}{2014}).
\newblock


\bibitem[\protect\citeauthoryear{Langford and the}{Langford and the}{2008}]%
        {langford_2008}
\bibfield{author}{\bibinfo{person}{John Langford} {and} \bibinfo{person}{Tong
  the}.} \bibinfo{year}{2008}\natexlab{}.
\newblock \showarticletitle{The epoch-greedy algorithm for contextual
  multi-armed bandits}. In \bibinfo{booktitle}{\emph{NIPS}}.
\newblock


\bibitem[\protect\citeauthoryear{Li, Ahmed, Ravi, and Smola}{Li
  et~al\mbox{.}}{2014}]%
        {li_2014}
\bibfield{author}{\bibinfo{person}{A.~Q. Li}, \bibinfo{person}{A. Ahmed},
  \bibinfo{person}{S. Ravi}, {and} \bibinfo{person}{A.~J. Smola}.}
  \bibinfo{year}{2014}\natexlab{}.
\newblock \showarticletitle{Reducing the sampling complexity of topic models}.
  In \bibinfo{booktitle}{\emph{KDD}}.
\newblock


\bibitem[\protect\citeauthoryear{Lu, Hoi, and Wang}{Lu et~al\mbox{.}}{2013}]%
        {lu2013second_order_filtering}
\bibfield{author}{\bibinfo{person}{Jing Lu}, \bibinfo{person}{Steven C.~H.
  Hoi}, {and} \bibinfo{person}{Jialei Wang}.} \bibinfo{year}{2013}\natexlab{}.
\newblock \showarticletitle{Second Order Online Collaborative Filtering}. In
  \bibinfo{booktitle}{\emph{ACML}}. \bibinfo{pages}{325--340}.
\newblock


\bibitem[\protect\citeauthoryear{Luostarinen and Kohonen}{Luostarinen and
  Kohonen}{2013}]%
        {Luostarinen}
\bibfield{author}{\bibinfo{person}{T. Luostarinen} {and} \bibinfo{person}{O.
  Kohonen}.} \bibinfo{year}{2013}\natexlab{}.
\newblock \showarticletitle{Using Topic Models in Content-Based news
  Recommender Systems}. In \bibinfo{booktitle}{\emph{Proc. of the 19th Nordic
  Conf. of Computational Linguistics}}. \bibinfo{pages}{239--251}.
\newblock


\bibitem[\protect\citeauthoryear{Murphy}{Murphy}{2012}]%
        {kevin_murphy_book_online_learning}
\bibfield{author}{\bibinfo{person}{Kevin~P. Murphy}.}
  \bibinfo{year}{2012}\natexlab{}.
\newblock \showarticletitle{Online Learning and Stochastic Optimization}.
\newblock \bibinfo{journal}{\emph{Machine Learning: A Probabilistic
  Perspective}} (\bibinfo{year}{2012}), \bibinfo{pages}{264--270}.
\newblock


\bibitem[\protect\citeauthoryear{Radlinski, Kleinberg, and Joachims}{Radlinski
  et~al\mbox{.}}{2008}]%
        {radlinski_2008}
\bibfield{author}{\bibinfo{person}{F. Radlinski}, \bibinfo{person}{R.
  Kleinberg}, {and} \bibinfo{person}{T. Joachims}.}
  \bibinfo{year}{2008}\natexlab{}.
\newblock \showarticletitle{Learning Diverse Rankings with Multi-Armed
  Bandits}. In \bibinfo{booktitle}{\emph{ICML}}.
\newblock


\bibitem[\protect\citeauthoryear{Rendle}{Rendle}{2010}]%
        {rendle2010fm}
\bibfield{author}{\bibinfo{person}{Steffen Rendle}.}
  \bibinfo{year}{2010}\natexlab{}.
\newblock \showarticletitle{Factorization Machines}. In
  \bibinfo{booktitle}{\emph{ICDM}} \emph{(\bibinfo{series}{ICDM '10})}.
  \bibinfo{publisher}{IEEE Computer Society}, \bibinfo{address}{Washington, DC,
  USA}, \bibinfo{pages}{995--1000}.
\newblock
\showISBNx{978-0-7695-4256-0}
\urldef\tempurl%
\url{https://doi.org/10.1109/ICDM.2010.127}
\showDOI{\tempurl}


\bibitem[\protect\citeauthoryear{Rosenblatt}{Rosenblatt}{1958}]%
        {rosenblatt_1958}
\bibfield{author}{\bibinfo{person}{F. Rosenblatt}.}
  \bibinfo{year}{1958}\natexlab{}.
\newblock \showarticletitle{The Perceptron: A Probabilistic Model for
  Information Storage and Organization in the Brain}.
\newblock \bibinfo{journal}{\emph{Psychological Review}} \bibinfo{volume}{65},
  \bibinfo{number}{6} (\bibinfo{year}{1958}), \bibinfo{pages}{386--408}.
\newblock


\bibitem[\protect\citeauthoryear{Shalev-Shwartz}{Shalev-Shwartz}{2011}]%
        {shalev_shwartz_2011}
\bibfield{author}{\bibinfo{person}{Shai Shalev-Shwartz}.}
  \bibinfo{year}{2011}\natexlab{}.
\newblock \showarticletitle{Online Learning and Online Convex Optimization}.
\newblock \bibinfo{journal}{\emph{Foundations and Trends in Machine Learning}}
  \bibinfo{volume}{4}, \bibinfo{number}{2} (\bibinfo{year}{2011}),
  \bibinfo{pages}{107--194}.
\newblock


\bibitem[\protect\citeauthoryear{Smith-Miles}{Smith-Miles}{2008}]%
        {miles_2008}
\bibfield{author}{\bibinfo{person}{Kate~A. Smith-Miles}.}
  \bibinfo{year}{2008}\natexlab{}.
\newblock \showarticletitle{Cross-Disciplinary Perspectives on Meta-Learning
  for Algorithm Selection}.
\newblock \bibinfo{journal}{\emph{Comput. Surveys}} \bibinfo{volume}{41},
  \bibinfo{number}{1} (\bibinfo{year}{2008}).
\newblock


\bibitem[\protect\citeauthoryear{Sriram and Makhani}{Sriram and Makhani}{[n.
  d.]}]%
        {galene_engine}
\bibfield{author}{\bibinfo{person}{S. Sriram} {and} \bibinfo{person}{A.
  Makhani}.} \bibinfo{year}{[n. d.]}\natexlab{}.
\newblock \showarticletitle{LinkedIn's Galene Search engine, 2014,
  https://engineering.linkedin.com/search/did-you-mean-galene.}
\newblock


\bibitem[\protect\citeauthoryear{Thompson}{Thompson}{1933}]%
        {thompson_1933}
\bibfield{author}{\bibinfo{person}{W.~R. Thompson}.}
  \bibinfo{year}{1933}\natexlab{}.
\newblock \showarticletitle{On the likelihood that one unknown probability
  exceeds another in view of the evidence of two samples}.
\newblock \bibinfo{journal}{\emph{Biometrika}} \bibinfo{volume}{25},
  \bibinfo{number}{3-4} (\bibinfo{year}{1933}).
\newblock


\bibitem[\protect\citeauthoryear{Wang, Kulkarni, and Poor}{Wang
  et~al\mbox{.}}{2005}]%
        {wang_2005}
\bibfield{author}{\bibinfo{person}{Chih-Chun Wang}, \bibinfo{person}{Sanjeev~R.
  Kulkarni}, {and} \bibinfo{person}{H.~Vincent Poor}.}
  \bibinfo{year}{2005}\natexlab{}.
\newblock \showarticletitle{Bandit problems with side observations}.
\newblock \bibinfo{journal}{\emph{IEEE Trans. Automatic Control}}
  \bibinfo{volume}{50}, \bibinfo{number}{3} (\bibinfo{year}{2005}),
  \bibinfo{pages}{338--355}.
\newblock


\bibitem[\protect\citeauthoryear{Yin, Sun, Cui, Hu, and Chen}{Yin
  et~al\mbox{.}}{2013}]%
        {LCARS}
\bibfield{author}{\bibinfo{person}{H. Yin}, \bibinfo{person}{Y. Sun},
  \bibinfo{person}{B. Cui}, \bibinfo{person}{Z. Hu}, {and} \bibinfo{person}{L.
  Chen}.} \bibinfo{year}{2013}\natexlab{}.
\newblock \showarticletitle{LCARS: A Location-Content-Aware Recommender
  System}. In \bibinfo{booktitle}{\emph{ACM KDD}}.
\newblock


\bibitem[\protect\citeauthoryear{Zhang, Zhou, Ma, Chen, Zhang, and
  Agarwal}{Zhang et~al\mbox{.}}{2016}]%
        {zhang2016glmix}
\bibfield{author}{\bibinfo{person}{XianXing Zhang}, \bibinfo{person}{Yitong
  Zhou}, \bibinfo{person}{Yiming Ma}, \bibinfo{person}{Bee-Chung Chen},
  \bibinfo{person}{Liang Zhang}, {and} \bibinfo{person}{Deepak Agarwal}.}
  \bibinfo{year}{2016}\natexlab{}.
\newblock \showarticletitle{GLMix: Generalized Linear Mixed Models For
  Large-Scale Response Prediction}. In \bibinfo{booktitle}{\emph{ACM KDD}}.
  \bibinfo{pages}{363--372}.
\newblock


\bibitem[\protect\citeauthoryear{Zhang, S{\'{e}}aghdha, Quercia, and
  Jambor}{Zhang et~al\mbox{.}}{2012}]%
        {Auralist}
\bibfield{author}{\bibinfo{person}{Y. Zhang}, \bibinfo{person}{D.
  S{\'{e}}aghdha}, \bibinfo{person}{D Quercia}, {and} \bibinfo{person}{T.
  Jambor}.} \bibinfo{year}{2012}\natexlab{}.
\newblock \showarticletitle{Auralist: Introducing serendipity into music
  recommendation}. In \bibinfo{booktitle}{\emph{ACM WSDM}}.
\newblock


\end{thebibliography}

\end{document}